\documentclass[10pt,twocolumn,letterpaper]{article}

\usepackage{wacv-2026-author-kit-template/wacv}              
\usepackage{comment}
\usepackage[accsupp]{axessibility}

\definecolor{wacvblue}{rgb}{0.21,0.49,0.74}
\usepackage{graphicx}
\usepackage{amsmath}
\usepackage{amssymb}
\usepackage{booktabs}
\usepackage{multirow}
\usepackage{siunitx}
\usepackage{subcaption}
\usepackage{svg}
\svgpath{{./}}
\usepackage{float}
\usepackage{pifont}
\newcommand{\cmark}{\ding{51}}  
\newcommand{\xmark}{\ding{55}}  
\usepackage[pagebackref,breaklinks,colorlinks,allcolors=wacvblue]{hyperref}

\def\wacvPaperID{926} 
\def\confName{WACV}
\def\confYear{2026}

\title{Fully Unsupervised Self-debiasing of Text-to-Image Diffusion Models}

\author{
Korada Sri Vardhana \qquad
Shrikrishna Lolla \qquad
Soma Biswas\\[4pt]
Indian Institute of Science, Bengaluru\\
{\tt\small srivardhanak@iisc.ac.in} \quad
{\tt\small shrikrishnalolla@gmail.com} \quad
{\tt\small somabiswas@iisc.ac.in}
}

\begin{document}
\maketitle
\begin{abstract}
Text-to-image (T2I) diffusion models have achieved widespread success due to their ability to generate high-resolution, photorealistic images. These models are trained on large-scale datasets, like LAION-5B, often scraped from the internet. However, since this data contains numerous biases, the models inherently learn and reproduce them, resulting in stereotypical outputs. We introduce \textbf{SelfDebias}, a fully unsupervised test-time debiasing method applicable to any diffusion model that uses a UNet as its noise predictor. SelfDebias identifies semantic clusters in an image encoder's embedding space and uses these clusters to guide the diffusion process during inference, minimizing the KL divergence between the output distribution and the uniform distribution. Unlike supervised approaches, SelfDebias does not require human-annotated datasets or external classifiers trained for each generated concept. 
Instead, it is designed to automatically identify semantic modes. Extensive experiments show that SelfDebias generalizes across prompts and diffusion model architectures, including both conditional and unconditional models. 
It not only effectively debiases images along key demographic dimensions while maintaining the visual fidelity of the generated images, but also more abstract concepts for which identifying biases is also challenging.
\end{abstract}    
\section{Introduction}
\label{sec:intro}

Generative models \cite{kingma2013auto, goodfellow2020generative, ho2020denoising}, particularly T2I models \cite{rombach2022high}, have achieved huge success in the past few years due to their ability to generate high-resolution, photorealistic images that align well with human perceptions. These models are often trained on massive datasets \cite{schuhmann2022laion5b}, which are scraped from the internet. Since a significant amount of bias exists in internet-sourced content, these models unintentionally learn such biases.
Recent studies \cite{shi2025dissecting} have shown that these biases are not only learned by the models but are also reproduced and sometimes amplified in their outputs. Since diffusion models are now commonly used in real-world systems \cite{perera2023analyzing, rosenberg2024limitations} and for data augmentation in downstream tasks \cite{trabucco2023effective}, the bias in the model's outputs, if left unchecked, can lead to unfair outcomes and significant representational harms \cite{parihar2024balancing}.

A natural solution to prevent this bias in the model's output is to re-train the model on curated unbiased datasets. 
But this approach suffers from degradation of diversity and visual fidelity \cite{perera2023analyzing}, 
in addition to the cost required to curate such datasets and retrain the models.
As a result, several alternative debiasing strategies have been proposed \cite{gandikota2024unified, parihar2024balancing,jiangdebiasing,chuang2023debiasing}.
Majority of the existing supervised approaches primarily address societal bias like gender and race, which are well known and can be pre-defined. 
Thus, these specific biased concepts can be learnt, and further used for debiasing. 
But, supervised debiasing frameworks are quite restrictive for free-form image generation: 
(i) Learning each concept separately, usually by generating sufficient number of examples of each using the model can be prohibitively expensive when debiasing along multiple combinations of attributes \cite{gandikota2024unified,parihar2024balancing};
(ii) In several scenarios, manually specifying the bias is very difficult, if not impossible. For example, for abstract concepts like a peaceful scenery, concrete concepts like bags, etc., it is not clear what are the biases introduced!
Thus, it is imperative to develop unsupervised techniques which can automatically discover the inherent biases and self-correct itself to generate more unbiased outputs.   
This is an extremely challenging and relatively unexplored direction. 
Recently, unsupervised frameworks such as TIME~\cite{orgad2023editing} and UCE~\cite{gandikota2024unified} have been proposed. However, both methods rely on a predefined list of attributes and their combinations that are intended to be debiased. Unlike TIME, which is restricted to editing attributes with only two categories, UCE enables the simultaneous debiasing of multiple categorical attributes. Consequently, most of our comparisons in this paper are made with UCE. UCE modifies the influence a token exerts along attribute directions in a specified ratio, making prior knowledge of these attributes essential, which may not always hold true in practice.

Towards this goal, we propose {\em SelfDebias}, a fully unsupervised debiasing framework that operates directly in the semantic latent space ($h$-space) of diffusion models. 
Rather than relying on predefined bias axes, we use the CLIP \cite{radford2021learning} image encoder to uncover semantic latent modes of variation through unsupervised clustering of generated image embeddings. These clusters serve as \textit{proxy} semantic groups, and we guide the generation process to reduce the dominance of any single group. Although we use CLIP in our experiments, any image encoder capable of semantically separating images in its embedding space could be used. This utilization of semantic separability in the embedding space, which is already available, enables us to perform debiasing.

The proposed SelfDebias framework has three main modules: 
{\bf (i) semantic projection module: } Here, a large batch of images is generated using a conditional diffusion model based on a given prompt. 
Then a projection network is learnt, which maps the $h$-space activations from the UNet of these generated images to their semantic image embeddings;
The {\bf (ii) semantic mode discovery} module performs hierarchical clustering to automatically cluster the projected features and discover semantic modes in the representation space;
During inference time, the {\bf (iii) self-debias module} uses these learned modes to perform soft cluster assignment over the projected $h$-space activations.
The KL divergence between the predicted soft cluster distribution and a uniform prior acts as a differentiable surrogate for balancing the concepts. 
The resulting loss is back-propagated to edit the $h$-space representations mid-generation, and the updated activations are used in the subsequent denoising steps.
This process allows us to apply semantic guidance directly in $h$-space without modifying the model weights or relying on any external supervision, making our method lightweight and compatible with existing pretrained diffusion models.

\textbf{Terminology.}
In the LLM literature ~\cite{schick2021selfdiagnosisselfdebiasingproposalreducing}, “self-debiasing” often denotes reflective or prompt-based procedures that leverage a model’s own knowledge to critique or revise its outputs . In this paper, by \emph{SelfDebias} we mean a purely \emph{test-time, statistical} intervention on hidden activations (the $h$-space). Crucially, model weights (U-Net, VAE, text encoder) are \emph{never} updated.

Our key contributions are as follows: 
\begin{enumerate}
\item To the best of our knowledge, SelfDebias is the first \textbf{fully} unsupervised, test-time framework for debiasing diffusion models, without requiring attribute labels, external classifiers, or retraining.  
\item We demonstrate that our inference-time intervention-based framework generalizes across different demographic attributes like gender, race, age, etc. without any supervision.  
\item We showcase the debiasing capability of SelfDebias in scenarios where the appropriate debiasing axis is unclear or difficult to define, such as abstract prompts, which cannot be handled by supervised techniques. 
\end{enumerate}
\section{Related Works}
\label{sec:prevworks}

Diffusion-based models are now the state-of-the-art for high-fidelity image generation. 
However, a growing body of research has shown that these models are not free from societal biases, like gender-occupation stereotypes, race, and age, \cite{basu2023inspecting, luccioni2023stable,cho2023dall,seshadri2023bias, luccioni2023stable}. 
The existing debiasing techniques can broadly be categorized into two approaches, namely debiasing with or without requiring retraining. \\
\textbf{Debiasing with retraining.} Many prior works attempt to debias generative models by retraining from scratch, as seen in the progression of Stable Diffusion (e.g., SD 1.5 vs. SD 2.1), or through fine-tuning \cite{choi2020fair, teo2023fair, um2023fair}. Some approaches use a balanced reference dataset to perform bias correction. Choi et al. \cite{choi2020fair} proposed a density-ratio estimation technique for bias identification and reweighted the training data accordingly. 
More recently, in the context of diffusion models, Shen et al. \cite{shen2023finetuning} propose a distributional alignment loss that directly optimizes for fairness in the generated outputs. 
Few recent works ~\cite{huang2025debiasing,kim2024training,hou2024invdiffinvariantguidancebias} have also explored debiasing by either fine-tuning or retraining the model on carefully curated datasets. 
While these retraining-based approaches have shown promising results, they depend on labeled or curated datasets and involves a substantial computational overhead for the retraining procedure. In contrast, our method does not require any retraining, access to the original training data, or demographic labels, substantially lowering the cost and complexity of deployment in real-world settings.  \\
\textbf{Debiasing without retraining.} These methods aim to intervene during inference time to mitigate bias, as opposed to fine-tuning-based approaches. As a result, they are significantly more cost-efficient in terms of training. A recent work by Parihar et al. \cite{parihar2024balancing} proposes a method for debiasing diffusion models by editing the intermediate latent representations ($h$-space) of the U-Net during inference. Similarly \cite{kwon2022diffusion,li2024self} directly perform edits in $h$-space to control semantic attributes, but both of these require at least some form of demographic supervision or annotation.
Chuang et al. \cite{chuang2023debiasing} performs algebraic manipulations of biased directions in text embeddings to debias \textit{T2I} diffusion models. Jiang et al. \cite{jiangdebiasing} argue that since text-image alignment occurs primarily through cross-attention, updating these layers using parameter-efficient tuning (e.g., LoRA-style adaptation) enables controlled attribute steering. \cite{orgad2023editing,gandikota2024unified} follows same principles and targets the cross-attention layers to debias and erase at the same time.
A recent work \cite{Shi_2025_CVPR}  detects bias features embedded in diffusion models at neuron level and intervenes to prevent bias. However, all these approaches assume access to labeled data and demographic dimensions to supervise optimization. NoiseCLR \cite{dalva2024noiseclr} also discovers self-supervised directions, but it requires costly training in noise space, does not apply to unconditional generation, and its directions capture general semantic variations rather than bias-specific axes.
\textit{This limits their general applicability, especially in cases where it is difficult to define debiasing axes. In contrast, we propose a fully unsupervised self-debiasing framework, which is applicable for a much wider range of concepts.}
\section{Proposed Framework}
\label{sec:method}
\begin{figure*}[t]
    \centering
    \includegraphics[width=0.8\textwidth]{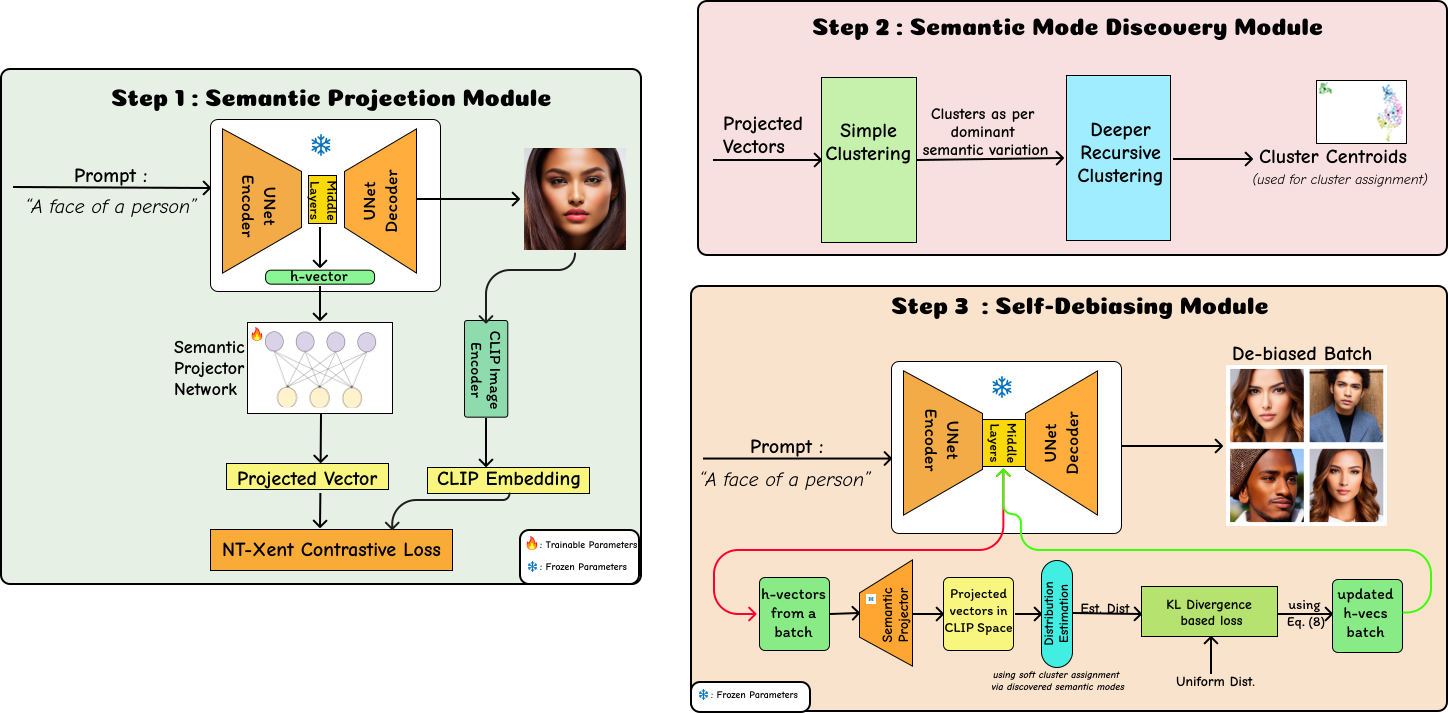}
    \caption{An overview of the proposed SelfDebias framework. 1) First, the projection network is trained to transform the semantically entangled {\em h-space} vectors to a disentangled semantic space; 2) Second, the semantic modes are discovered through a two stage clustering process of the projected \textit{h}-vectors; 3) Finally, the \textit{h}-vectors of a biased batch are updated such that the tend towards a more uniform distribution, leading to a debiased batch.}
    \label{fig:full-pipeline}
\end{figure*}

Here, we describe SelfDebias, a fully unsupervised debiasing framework for pretrained diffusion models, without retraining, supervision, or access to attribute labels. 
The key idea is to align the distribution of generated samples across latent semantic modes, discovered via clustering in the image encoder's embedding space, to a uniform target, thereby reducing the inherent bias. 
The framework consists of three key modules (Fig.~\ref{fig:full-pipeline}): 
(1) Semantic Projection Module: where a projection from \textit{h-space} to semantic space is learnt; 
(2) Semantic Mode Discovery Module: where semantic clusters are determined via unsupervised clustering; and 
(3) Self-debias Module: where \textit{h-space} editing via a KL-divergence loss is performed between the predicted cluster distribution and a uniform prior.
We will now describe these modules.

\subsection{Semantic Projection Module}
\label{sec:projection}
Here, we learn a projection from the \textit{h-space} to a semantically disentangled latent space. \\
{\bf Intermediate Activations ({\em h-space}):} 
We briefly describe the \textit{h-space} of diffusion models, which is the intermediate representation for completion. 
The U-Net backbone used in diffusion models has a bottleneck feature layer between the encoder and decoder blocks. The activations at this bottleneck layer are referred to as \textit{h-space}~\cite{kwon2022diffusion}, denoted \( h_t \). These intermediate representations contain rich semantic information about the current sample at timestep \( t \), and have been shown to encode interpretable features such as gender, age, and object type \cite{kwon2022diffusion}. 
In our framework, we extract \( h_t \) during the sampling process and apply gradient-based updates 
This makes our method efficient, modular, and applicable to any diffusion model with U-Net-style architectures.

\begin{figure}[t!]
    \centering
    \includegraphics[width=1\linewidth]{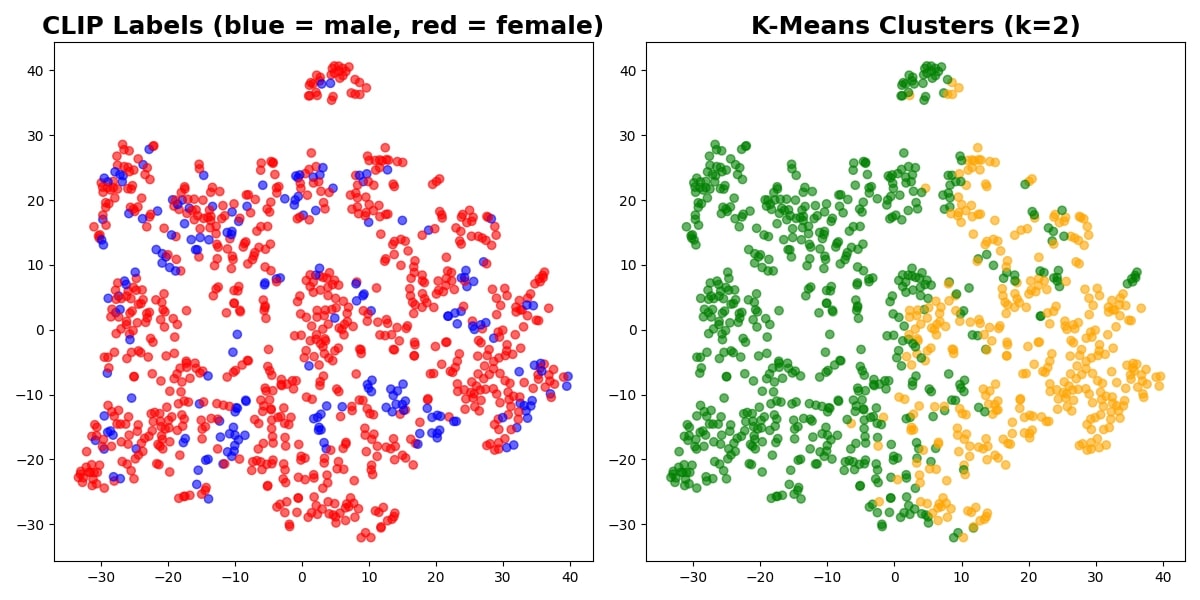}
  \caption{
    Left: \textit{h}-space vectors labeled using CLIP as a zero-shot classifier (blue = male, red = female).
    Right: The same \textit{h-vectors} grouped via K-Means clustering ($k=2$).
    Both visualizations are produced using PCA followed by t-SNE.
    The unsupervised clustering (right) fails to align meaningfully with the semantic labels from CLIP (left). This misalignment highlights the entangled nature of the \textit{h-space}.
    }
    \label{fig:h_space}
\end{figure}

Recent supervised debiasing framework \cite{parihar2024balancing} has utilized \textit{h-space}. First, we perform a simple analysis on the \textit{h-space} to understand whether they are sufficiently discriminative and can be directly clustered to determine the underlying concepts. We take gender (male and female) as the running example, for ease of understanding. 
We generated 2,000 images using the prompt \textit{"A face of a person"} and capture their \textit{h}-space embeddings during generation. These embeddings are clustered and visualized, and the corresponding images are also labeled using CLIP as a zero-shot classifier. Both visualizations are shown side by side in Fig.~\ref{fig:h_space}. 
 We observe that although these features are semantically rich, they are also highly entangled. 
This prevents effective clustering in this space for the purpose of extracting distinct semantic modes. 
Therefore, we need to project the $h$-space vectors into an embedding space where the representations are more semantically disentangled.

Let \( h_t^{(i)} \in \mathbb{R}^d \) denote the $h$-space activation at timestep \( t \) for image \( i \), extracted from the bottleneck layer of the U-Net. Let \( x_0^{(i)} \) be the corresponding generated image obtained via decoding, and let \( z^{(i)} = E(x_0^{(i)})\) be its embedding in the semantic space, where \( E \) is the image encoder (we used CLIP), capable of encoding images into a semantically disentangled space. We train a lightweight projection network \( g_\psi \) to align $h$-space activations with their corresponding semantic embeddings. To achieve this, we use the NT-Xent contrastive loss \cite{chen2020simple}, which pulls together positive pairs \( (g_\psi(h_t^{(i)}), z^{(i)}) \) while pushing apart negative pairs \( (g_\psi(h_t^{(i)}), z^{(j)}) \) for \( j \neq i \) within the same batch.
For a batch of size \( N \), the loss for a sample \( i \) is defined as:
\begin{equation}
\mathcal{L}_{\text{proj}}^{(i)} = -\log \frac{\exp\left(\cos(g_\psi(h_t^{(i)}), z^{(i)}) / \tau\right)}{\sum_{j=1}^N \exp\left(\cos(g_\psi(h_t^{(i)}), z^{(j)}) / \tau\right)},
\end{equation}
where \( \cos(\cdot, \cdot) \) denotes cosine similarity and \( \tau \) is the temperature hyperparameter.
The total projection loss over the batch is given by:
\begin{equation}
\mathcal{L}_{\text{proj}} = \frac{1}{N} \sum_{i=1}^N \mathcal{L}_{\text{proj}}^{(i)}.
\end{equation}
This encourages the projected $h$-space features to align with their corresponding semantic embeddings, while remaining well-separated from others in the batch. Once trained, \( g_\psi \) is frozen and used during inference to project $h$-space activations into the semantic embedding space.

\subsection{ Semantic Mode Discovery Module}
\label{sec:clustering}

After semantic projection, we obtain a set of projected semantic embeddings \( \{\hat{z}_t^{(i)}\}_{i=1}^{N} \) of \( N \) images, where \( \hat{z}_t^{(i)} = g_\psi(h_t^{(i)}) \). To uncover latent semantic modes of variation without supervision, we perform clustering on the projected embeddings. We do not assume prior knowledge of the number of clusters, as the relevant modes of variation may not be obvious. \\ 
{\bf Challenge in discovering modes:} To better understand what kind of clustering is required to compute the different concepts, we again consider the widely used demographic concepts, but here, instead of only gender, we also consider other factors like race, age, etc. We observe  that though gender is well clustered, the other factors are quite entangled. 
As expected, we cannot disentangle across multiple dimensions like race and gender simultaneously, since each gender will have people of all races. With more dimensions, the entanglement further increases. 
To discover all these factors in an unsupervised manner is extremely challenging.
Here, we propose to use a hierarchical clustering \cite{roux2018comparative} with adaptive weighting to discover these semantic models to eventually debias the model output. \\
{\bf First Stage Clustering:} To estimate the appropriate number of clusters, we evaluate clustering quality across different $k$ values using silhouette score \cite{9260048}, which quantifies how well each point fits within its assigned cluster relative to other clusters. For each $k$, we compute the average silhouette score across all data points. Higher score indicates that the clusters are well-separated and internally cohesive. We select the $k$ value closest to the maximum silhouette score. This aligns with the usual model selection principles in unsupervised learning. \\
{\bf Second Stage Clustering:} The clusters obtained up to this stage capture the dominant semantic groupings in the data, i.e., those aligned along axes of highest semantic variation. However, each cluster may still contain finer-grained semantic distinctions. 
We apply recursive spectral clustering within each to uncover finer semantic dimensions.
This recursive process is governed by two hyperparameters: \( s_{\min} \) (minimum allowed cluster size) and \( d_{\max} \) (maximum recursion depth). A cluster is split only if it contains more than \( s_{\min} \) samples and its current depth \( d < d_{\max} \).
This results in a hierarchy of clusters with varying semantic granularity. We denote the final set of leaf clusters as \( \mathcal{C} = \{c_1, c_2, \dots, c_K\} \), where each cluster \( c_j \) is associated with a semantic embedding space centroid and a depth \( d_j \in \{0, 1, \dots, d_{\max}\} \).
For each projected vector \( \hat{z}_t^{(i)} \), we compute a soft assignment over the final set of clusters using cosine similarity with temperature scaling:
\begin{equation}
p_t^{(i)}[j] = \frac{\exp\left(\alpha \cdot \cos(\hat{z}_t^{(i)}, c_j)\right)}{\sum_{k'=1}^{K} \exp\left(\alpha \cdot \cos(\hat{z}_t^{(i)}, c_{k'})\right)}
\end{equation}
where \( \alpha \) is a temperature parameter controlling assignment sharpness.
We then average these soft assignments across the images to obtain the empirical distribution over discovered clusters as follows:
\begin{equation}
P_t[j] = \frac{1}{N} \sum_{i=1}^{N} p_t^{(i)}[j]
\end{equation}
{\bf Adaptive Cluster Weightage:} To reflect the hierarchical structure of the clusters, we assign each cluster \( c_j \) a weight based on its depth:
\begin{equation}
w_j = \frac{1}{d_j + 1}
\end{equation}
The target distribution is then defined as:
\begin{equation}
\mathcal{U}[j] = \frac{w_j}{\sum_{k=1}^{K} w_k}
\end{equation}
This depth-weighted uniform target ensures that shallower semantic modes are treated as more significant, while allowing fine-grained sub-clusters to contribute proportionally. 
\subsection{Self-debiasing Module}
\label{sec:kl_update}

To encourage balanced representation of the discovered latent modes, inspired by \cite{parihar2024balancing}, we minimize the KL divergence between the predicted distribution \( P_t \) and the depth-weighted uniform target \( \mathcal{U} \):
\begin{equation}
\mathcal{L}_{\text{uniform}} = \text{KL}(P_t \,\|\, \mathcal{U}) = \sum_{j=1}^{K} P_t[j] \log \left( \frac{P_t[j]}{\mathcal{U}[j]} \right)
\end{equation}
{\em Our novel formulation facilitates debiasing across all the axes (gender, age, race) simultaneously.
Thus the target distribution will have all the concepts that have been discovered in the previous module.}
Since the assignment probabilities are defined via a softmax over cosine similarities, they form a differentiable function of the projected \textit{h}-vectors, making the loss in Eq.~(7) differentiable.
We back-propagate this loss into the {\em h-space} representations:
\begin{equation}
\tilde{h}_t^{(i)} = h_t^{(i)} - \gamma \cdot \nabla_{h_t^{(i)}} \mathcal{L}_{\text{uniform}}
\end{equation}
The updated \( \tilde{h}_t^{(i)} \) is then used to compute the denoised latent \( x_{t-1}^{(i)} \) in the sampling trajectory.
Notably, the entire process is unsupervised and test-time only, requiring no retraining, no attribute labels, and no access to the training dataset. \\
{\bf Overall Framework:} To summarize, we begin by generating a set of 2,000 images along with their corresponding \textit{h}-space activations for a given prompt. These activations are then used to train the semantic projection module. Once trained, this module is used to project the \textit{h}-space vectors into the semantic embedding space, where clustering is performed. After identifying the cluster centroids, we intervene in the \textit{h}-space during inference, using the method described above, to steer each generated image batch towards more balanced outputs.
\section{Experiments}
\label{sec:experiments}

Here, we present extensive experiments to evaluate the efficacy of the proposed framework. 
First, we evaluate the effectiveness of our approach in debiasing the face domain, addressing both major semantic shifts and finer-grained variations, and compare it against existing methods.
Additionally, we assess SelfDebias on occupational bias using prompts derived from the WinoBias~\cite{zhao2018gender} dataset. 
Notably, we highlight SelfDebias’s unique ability to debias even when the target attributes are vague or unspecified. We further demonstrate the method’s independence from any specific image encoder and its model-agnostic nature.
    
\subsection{Implementation Details}
For our experiments, we use Stable Diffusion v1.5~\cite{rombach2022high}. 
All experiments are conducted on NVIDIA RTX A5000 GPUs. We used a batch size of 100 for dataset generation and performed image synthesis with 50 inference steps. For training the semantic projection module, we use a time-conditioned two-layer MLP  projecting to 512 dimensions. Its input is a flattened $h$-vector and a timestep embedding and it outputs a 512-dimensional vector aligned with the CLIP space.
For evaluation, we use Fairness Discrepancy (FD)~\cite{choi2020fair} to quantify demographic bias, which measures the deviation of the generated attribute distribution from a uniform target. FD is computed as:
\begin{equation}
\mathrm{FD} = \left\| \bar{p} - \mathbb{E}_{x \sim p_\theta(x)}[C(x)] \right\|_2,
\end{equation}
where $p_\theta(x)$ is the distribution of generated images and $\bar{p}$ denotes the uniform distribution over attribute classes. $C(\cdot)$ is a high-accuracy attribute classifier, for which we have used same setup as in \cite{parihar2024balancing}. 
We used FID \cite{heusel2017gans} to determine the quality of the images generated. We have used standard implementation of clean-fid \cite{parmar2022aliased} for calculating it.

\noindent
{\bf Baseline Approaches:} 
To the best of our knowledge, there are no fully unsupervised techniques, so we compare with the following approaches:
1) {\bf Supervised Approaches:} Here, we compare with i) Random Sampling, which is vanilla stable diffusion model, without any modifications. ii) ITI-Gen \cite{zhang2023iti} , which works through appending of prompt tokens with learned prompt embeddings of each category. iii) Fair Diffusion \cite{friedrich2023fair}, which uses a lookup table for recognizing bias from the given text input and adds corresponding scaled embeddings to the text embeddings  iv) H-Guidance \cite{parihar2024balancing}, which not only assumes the knowledge of the bias concepts, but also learns the specific concepts by generating sufficient images using the base diffusion model; 
2) {\bf Unsupervised Approach with Bias Knowledge}: We also compare with the recent unsupervised approach UCE \cite{gandikota2024unified} and self-supervised approach ID \cite{li2024self}, both of which assume the knowledge of the bias concepts for performing debiasing, but doesn't use any labeled datasets to achieve their debiasing objectives. Unlike our method, ID \cite{li2024self} learns a separate direction for each attribute; for fairness, we selected concepts via a uniform-probability switch rather than manual toggling.

\subsection{Mitigating Bias in Human Face Generation}
\label{sec:main-clustering}
Here, we present the results of our approach using (i) two clusters and (ii) multiple clusters.  \\
{\bf Two Cluster Results:} 
We observe that the image distributions produced by diffusion models naturally separate into a few dominant semantic clusters in CLIP embedding space, which typically align with high-variance attributes such as gender in faces or species in pets. 
In the first experiment, we cluster the embeddings of human faces into two clusters (which naturally align with gender) and report the performance in Table~\ref{tab:face-2}.
The results of the competing methods are taken directly from Parihar et al. \cite{parihar2024balancing} for consistency and comparability.
We observe that despite being a completely unsupervised approach, it outperforms supervised techniques in terms of FD and is third in terms of FID (only slightly worse than UCE).

\begin{table}[t!]
\centering
\caption{Comparison of different methods for debiasing gender using FD and FID metrics. We also indicate whether the method assumes known bias labels or operates in an unsupervised manner.
\underline{Underline} = Best overall, \textbf{bold} = Best among unsupervised methods;
Note that ours is completely unsupervised, whereas UCE and ID assume the knowledge of the bias axes.}
\label{tab:results}
\resizebox{\columnwidth}{!}{%
\begin{tabular}{
    l
    S[table-format=1.3]  
    S[table-format=2.2]
    c
    c
}
\toprule
\textbf{Method} & \textbf{FD$\downarrow$} & \textbf{FID$\downarrow$} & \textbf{Bias unknown?} & \textbf{No supervision?} \\
\midrule
Random Sampling                                & 0.317 & 72.37 & --- & --- \\
ITI-Gen~\cite{zhang2023iti}            & 0.049 & \underline{64.79} & \xmark & \xmark \\
Fair Diffusion~\cite{friedrich2023fair}& 0.227 & 71.22 & \xmark & \xmark \\
H-Guidance~\cite{parihar2024balancing} & 0.024 & 70.69 & \xmark & \xmark \\
\hline
UCE~\cite{gandikota2024unified}        & 0.028 & \textbf{69.81} & \xmark & \color{green} \cmark \color{black} \\
ID~\cite{li2024self} & 0.019 & 70.64 & \xmark & \color{green} \cmark \color{black} \\
\textbf{SelfDebias (ours)}                   & \underline{\textbf{0.009}} \color{green} & 70.52 & \color{green}\cmark\color{black} & \color{green} \cmark \color{black} \\
\bottomrule
\end{tabular}
}
\label{tab:face-2}
\end{table}

\begin{figure}
    \centering
    \includegraphics[width=1\linewidth]{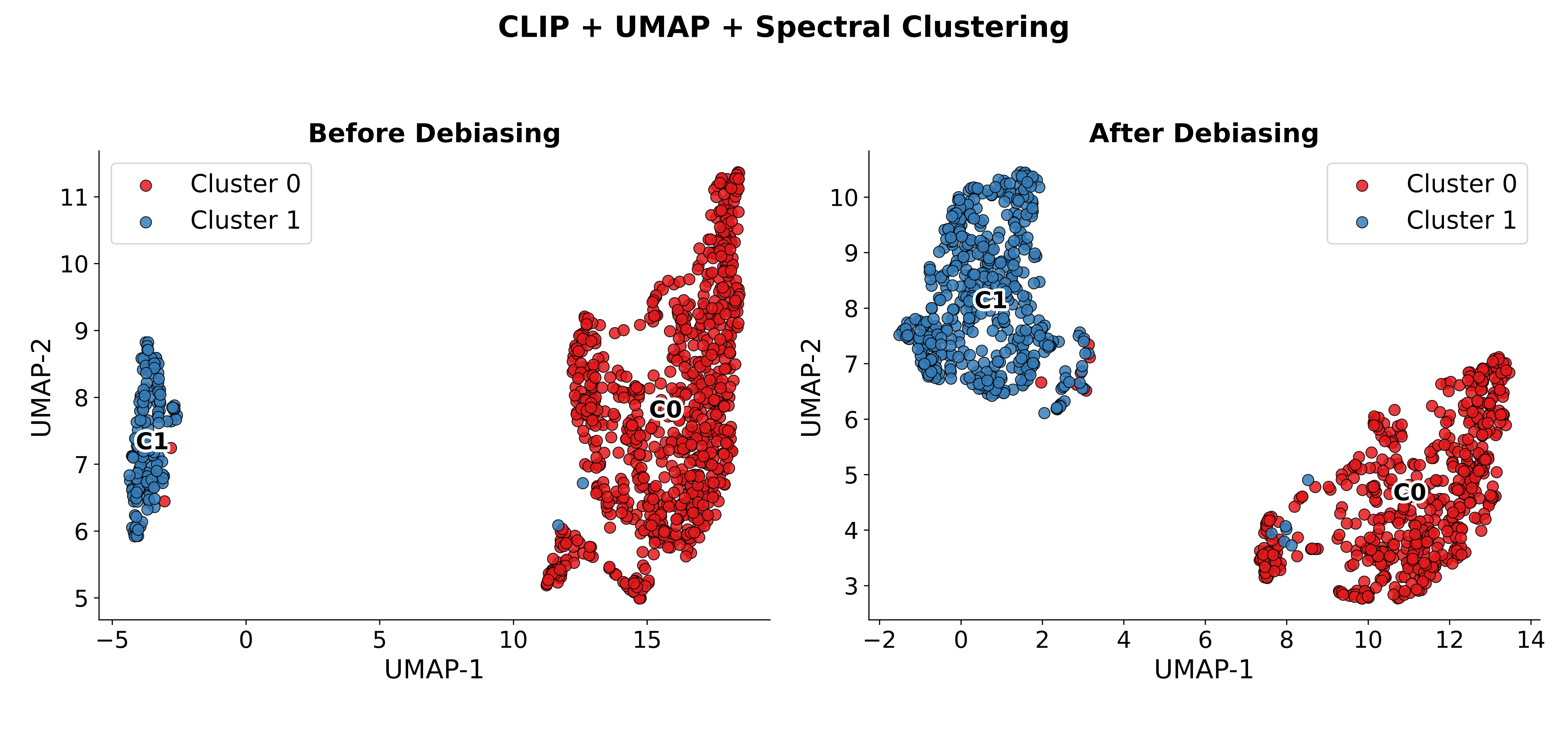}
    \caption{UMAP visualization of image embeddings before (left) and after (right) debiasing for gender. Right plot exhibits more balanced cluster distribution, indicating effective mitigation of bias.}
    \label{fig:simple_faces_umap}
\end{figure}

\begin{figure}
    \centering
    \includegraphics[width=0.8\linewidth]{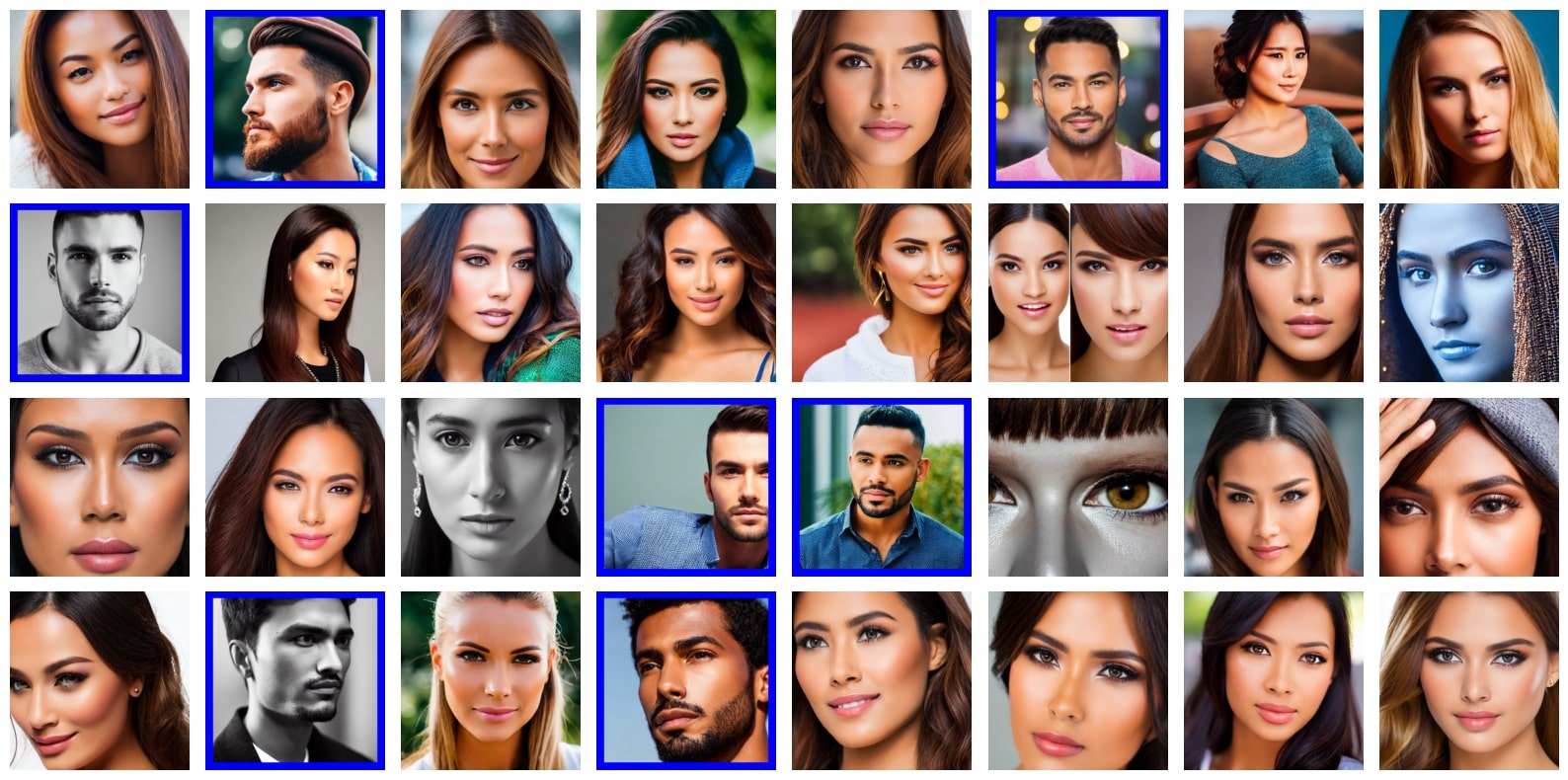}
    \par\vskip 0.5em
    \includegraphics[width=0.8\linewidth]{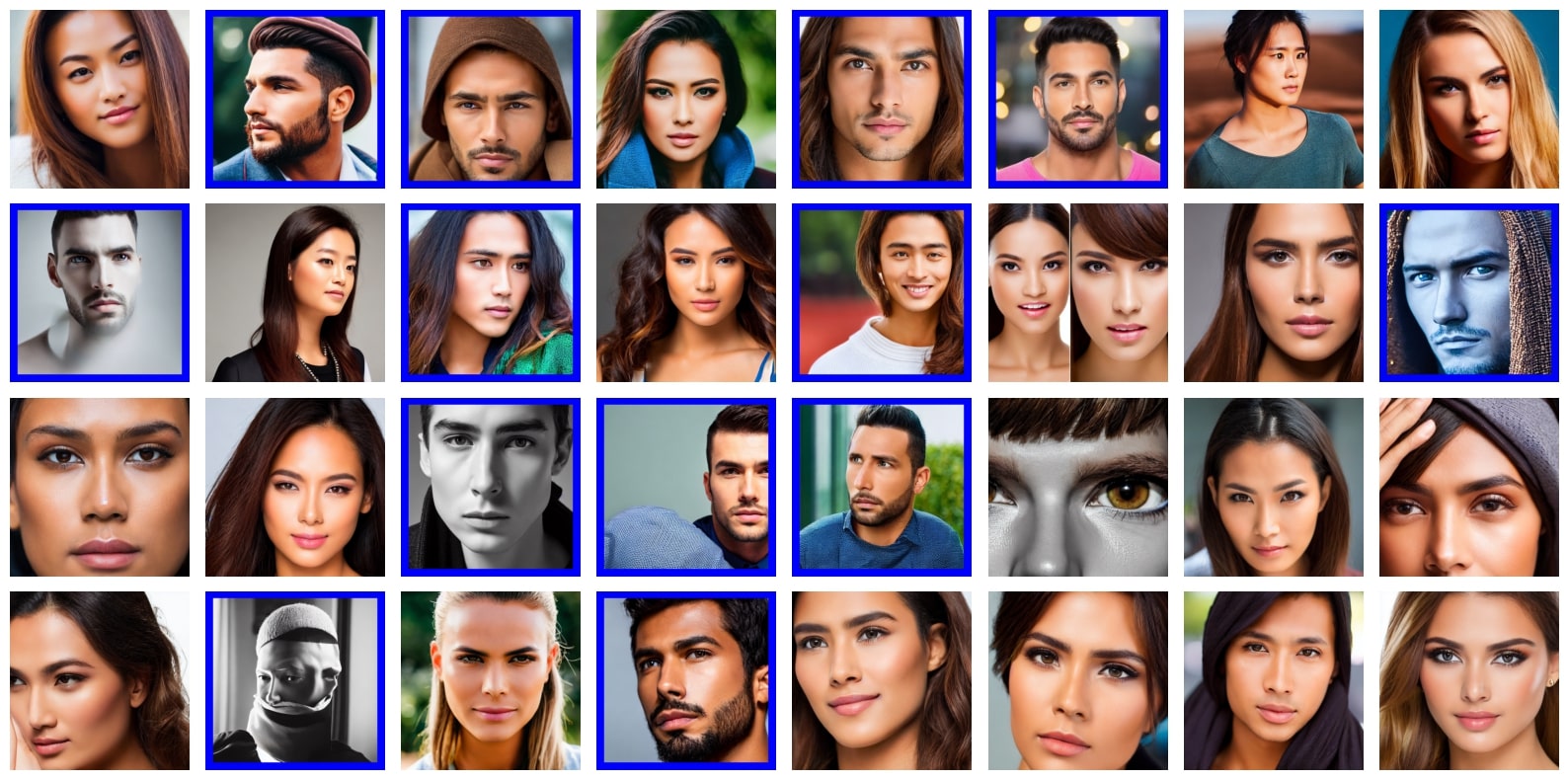}
    \caption{
        Top: Generated images using vanilla Stable Diffusion 1.5. \color{black}
        Bottom: Generated images using SelfDebias. The debiased set contains 44\% men, as opposed to 22\% using the original model (in blue).
    }
    \label{fig:stacked_faces}
\end{figure}

For qualitative analysis, we present visualizations of images in CLIP embedding space before and after debiasing in Fig.~\ref{fig:simple_faces_umap} and image grids before and after debiasing in Fig.~\ref{fig:stacked_faces}. Together, these visualizations illustrate how our method restructures the semantic space and promotes greater balance across dominant demographics. \\
{\bf Multiple Clusters:} Here, we further divide each main cluster found in above clustering for debiasing along more fine-grained attributes, like age and race. Despite the fact that these sub-clusters cannot simultaneously align cleanly with all attributes, they often represent consistent local semantics, what we call \textit{pseudo-semantic regions}. We observe that it captures local structure that reflects entangled but meaningful groupings.

We identify such regions recursively and use their centroids to guide debiasing. At inference time, we compute similarity of latents to these centroids and enforce a form of semantic diversity by encouraging a uniform sampling over them. Importantly, we do not attempt to identify each attribute axis explicitly, nor to enumerate all possible demographic intersections, both of which can be combinatorially large and often unobservable. Instead, we rely on the emergent structure revealed through recursive unsupervised clustering. Though the number of resulting clusters is often fewer than the total number of demographic combinations, we empirically find that enforcing balance across these pseudo-semantic regions reduces demographic bias, as measured by the evaluation metrics.

\begin{figure}
    \centering
    \includegraphics[width=0.8\linewidth]{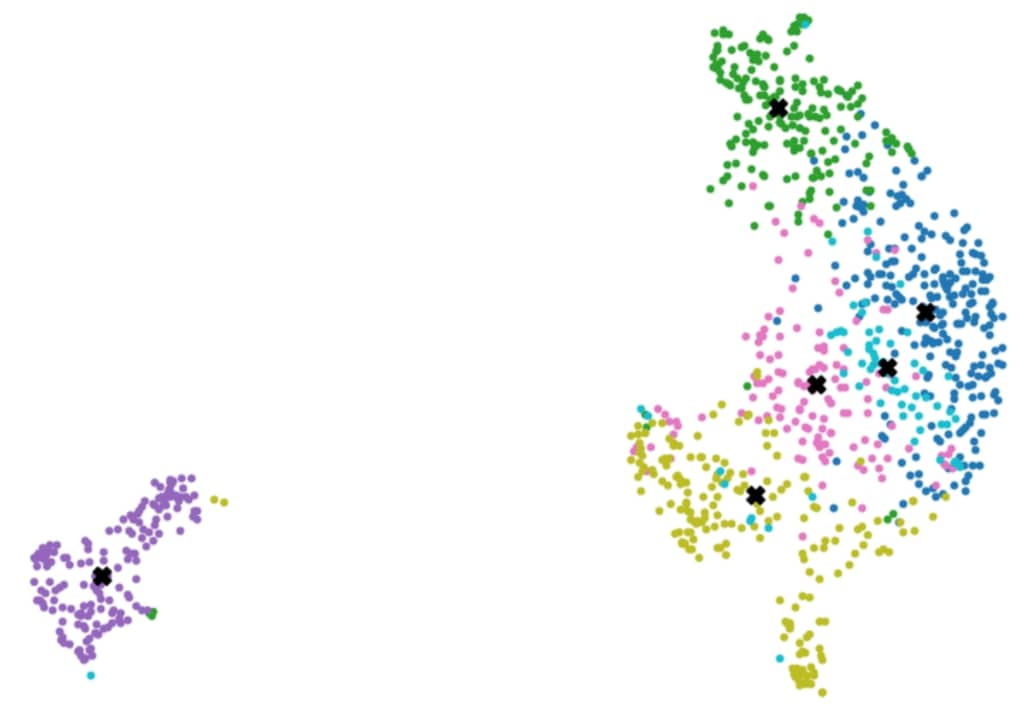}
    \caption{UMAP projection showing fine-grained clusters and their centroids (marked with black-crosses) for faces generation.}
    \label{fig:recursive_umap}
\end{figure}

\begin{figure}
    \centering
    \includegraphics[width=0.8\linewidth]{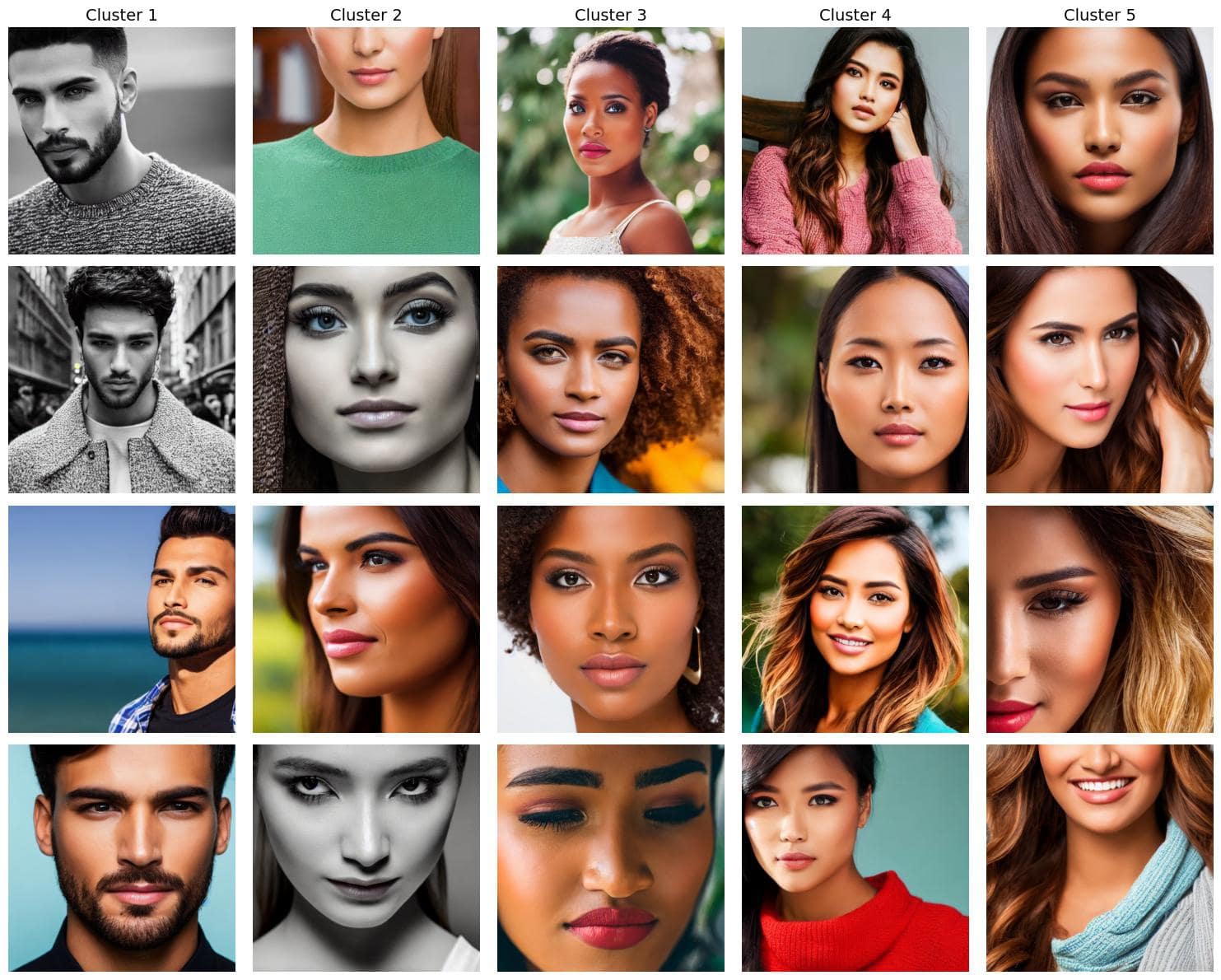}
    \caption{Results of hierarchical clustering; each column displays a subset of images from one cluster. Column 1 primarily contains men. Cluster 3 includes individuals with darker skin tones, while Cluster 4 predominantly features individuals with Asian facial features. Clusters 2 and 5 appear more semantically mixed, including faces with characteristics commonly associated with Indian and white individuals.}

    \label{fig:rand_grid}
\end{figure}

As shown in Table~\ref{tab:rec_fd}, recursive debiasing improves parity across gender, race and their combinations. 
Fig.~\ref{fig:recursive_umap} visualizes the discovered cluster hierarchy using UMAP, illustrating how deeper recursive levels uncover nuanced structure otherwise hidden in flat representations. We show sample images from each cluster in Fig. ~\ref{fig:rand_grid}.
We observe that SelfDebias outperforms even the supervised approaches for gender and Race+Age+Gender criteria, despite being a completely unsupervised approach, unlike the other compared approaches. \textit{For these results, the supervised techniques generate images for each concept at a time, for example, for age, images are generated specifically for diverse ages, whereas for UCE \cite{gandikota2024unified}, we enumerate all the concepts. Our framework debiases across all the dimensions at once.}

\subsection{Occupation-Related Bias}
 Next, we evaluate SelfDebias on a set of occupation-based prompts inspired by the Winobias benchmark (e.g., ``a photo of a doctor'', ``a photo of a teacher''), which are known to elicit biased gender associations in generative models due to correlations in web-scale training data. For each prompt, we generate a batch of images using our method and then classify the gender of each generated image to compute the deviation ratio \cite{gandikota2024unified} for evaluation.
Importantly, we perform debiasing using cluster centroids computed from a single generic prompt: ``a photo of a person''. This decouples the clustering step from the downstream prompts and demonstrates that SelfDebias does not require task-specific representations at test time. As shown in Table~\ref{tab:winobias_table}, our method consistently reduces gender skew, even though the semantic clusters were learned using a neutral base prompt. We observe that for all three of the four prompts, it outperforms UCE, whereas for Analyst, it performs almost at par with the others. 
The results for other methods in Table \ref{tab:winobias_table} are directly taken from \cite{gandikota2024unified}  for consistency in comparison.

\begin{table}[t]
\centering
\caption{
Comparison of different methods for debiasing using the FD and FID metrics.  
CNA = Checkpoint Not Available.  
Criteria RAG = Race + Age + Gender.  
Race includes White, Black, Indian \& Asian. Age includes Young, Old \& Adult.  
\underline{Underline} = Best overall, \textbf{bold} = Best among unsupervised methods, * = Uses supervision, \$ = Unsupervised but knows labels
}
\label{tab:rec_fd}
\resizebox{\columnwidth}{!}{%
\begin{tabular}{l
                cc  
                cc  
                cc  
                cc} 
\toprule
\textbf{Criteria} 
& \multicolumn{2}{c}{\textbf{Random Sampling}} 
& \multicolumn{2}{c}{\textbf{H-Guid\textsuperscript{*}~\cite{parihar2024balancing}}} 
& \multicolumn{2}{c}{\textbf{UCE\textsuperscript{\$}~\cite{gandikota2024unified}}} 
& \multicolumn{2}{c}{\textbf{SelfDebias (Ours)}} \\
\cmidrule(lr){2-3}
\cmidrule(lr){4-5}
\cmidrule(lr){6-7}
\cmidrule(lr){8-9}
& FD & FID & FD & FID & FD & FID & FD & FID \\
\midrule
Race   & 0.387 & 72.37 & \underline{0.188} & 75.52 & 0.254 & 83.84 & \textbf{0.237} & 87.08 \\
Age    & 0.488 & 72.37 & \underline{0.194} & 78.88 & \textbf{0.238} & 87.82 & 0.272 & 87.08 \\
Gender & 0.317 & 72.37 & 0.024 & 70.69 & 0.028 & 69.81 & \underline{\textbf{0.015}} & 87.08 \\
RAG    & 0.385 & 72.37 & CNA & CNA & 0.314 & 89.96 & \textbf{0.297} & 87.08 \\
\bottomrule
\end{tabular}
}
\end{table}

\begin{table*}[t]
\centering
\small
\caption{Profession-specific gender deviation ratios (mean $\pm$ std) across different debiasing methods. Lower is better.\underline{Underline} = Best overall, \textbf{bold} = Best among unsupervised methods}
\resizebox{\textwidth}{!}{%
\begin{tabular}{lccccccc}
\toprule
\textbf{Profession} & \textbf{Original-SD} & \textbf{Concept Algebra}~\cite{wang2023concept} & \textbf{Debias-VL}~\cite{chuang2023debiasing} & \textbf{TIME}~\cite{orgad2023editing} & \textbf{TIME + Preserve}~\cite{gandikota2024unified} & \textbf{UCE}~\cite{gandikota2024unified} & \textbf{SelfDebias} \\
\midrule
Librarian & $0.86 \pm 0.06$ & $0.66 \pm 0.07$ & $0.34 \pm 0.06$ & $0.26 \pm 0.05$ & $0.35 \pm 0.01$ & $0.07 \pm 0.07$ & \underline{$\mathbf{0.06 \pm 0.05}$} \\
Teacher   & $0.42 \pm 0.01$ & $0.46 \pm 0.00$ & $0.11 \pm 0.05$ & $0.34 \pm 0.06$ & $0.07 \pm 0.06$ & $0.06 \pm 0.02$ & \underline{$\mathbf{0.04 \pm 0.02}$} \\
Analyst & $0.58 \pm 0.12$ & $0.24 \pm 0.18$ & $0.71 \pm 0.02$ & $0.52 \pm 0.03$ & \underline{$\mathbf{0.13 \pm 0.05}$} & $0.20 \pm 0.07$ & $0.22 \pm 0.06$ \\
Doctor    & $0.78 \pm 0.04$ & $0.40 \pm 0.02$ & $0.50 \pm 0.04$ & $0.58 \pm 0.03$ & $0.41 \pm 0.08$ & $0.20 \pm 0.02$ & \underline{$\mathbf{0.18 \pm 0.04}$} \\
\bottomrule
\end{tabular}
}
\label{tab:winobias_table}
\end{table*}

\subsection{Debiasing Abstract Concepts}
\begin{figure}
    \centering
    \includegraphics[width=0.8\linewidth]{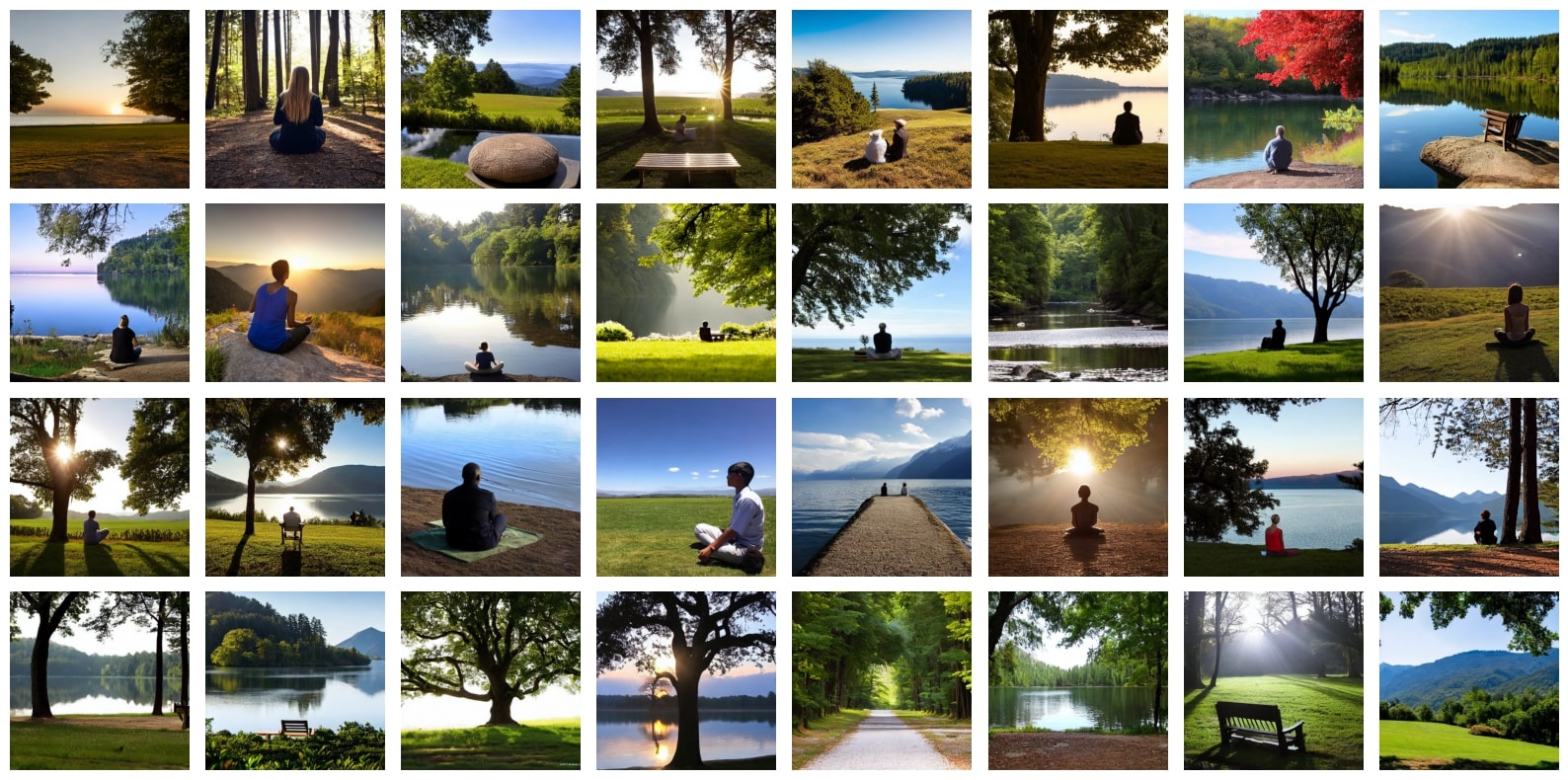}
    \par\vskip 0.5em  
    \includegraphics[width=0.8\linewidth]{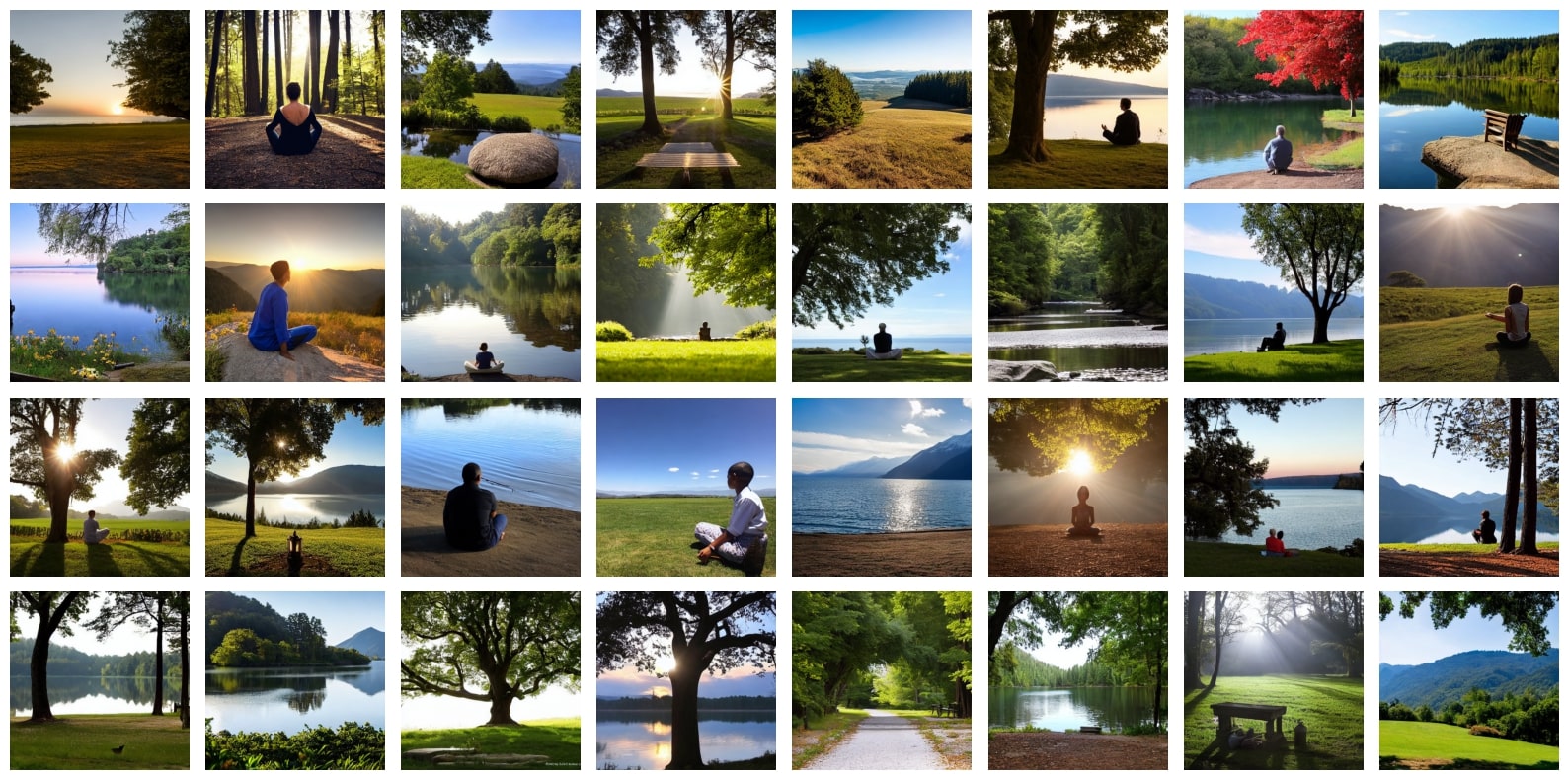}
    \caption{
        Top: Generated images using original model Bottom: Generated images after applying SelfDebias. In the original set (top), 20 out of 32 images (62.5\%) depict scenes without people. After debiasing (bottom), this number decreases to 16 out of 32 images (50\%), reflecting a 12.5 \% reduction in people-less scenes.
    }
    \label{fig:natural_env_grid}
\end{figure}

The proposed SelfDebias is unique in its ability to perform debiasing without any prior knowledge of the bias concepts.
This enables our method to generalize beyond predefined notions of bias and operate effectively in settings where the source of bias is unknown or unlabelled.
To demonstrate our method's ability to operate without prior bias axes, we present two illustrative examples. In the first case, we prompt a Stable Diffusion model with the phrase \textit{"a peaceful moment"}. The model typically generates either empty scenic landscapes or landscapes with people, often favoring the former disproportionately. In this context, it is not feasible to define an explicit bias axis, as the notion of “peacefulness” is abstract and subjective. Despite the lack of a clearly identifiable bias direction, SelfDebias effectively identifies and mitigates the over-representation of people in the outputs. As shown in Fig.~\ref{fig:natural_env_grid}, SelfDebias successfully debiases such ambiguous scenarios without relying on explicit guidance or external classifiers.

\begin{figure}
    \centering
    \includegraphics[width=0.8\linewidth]{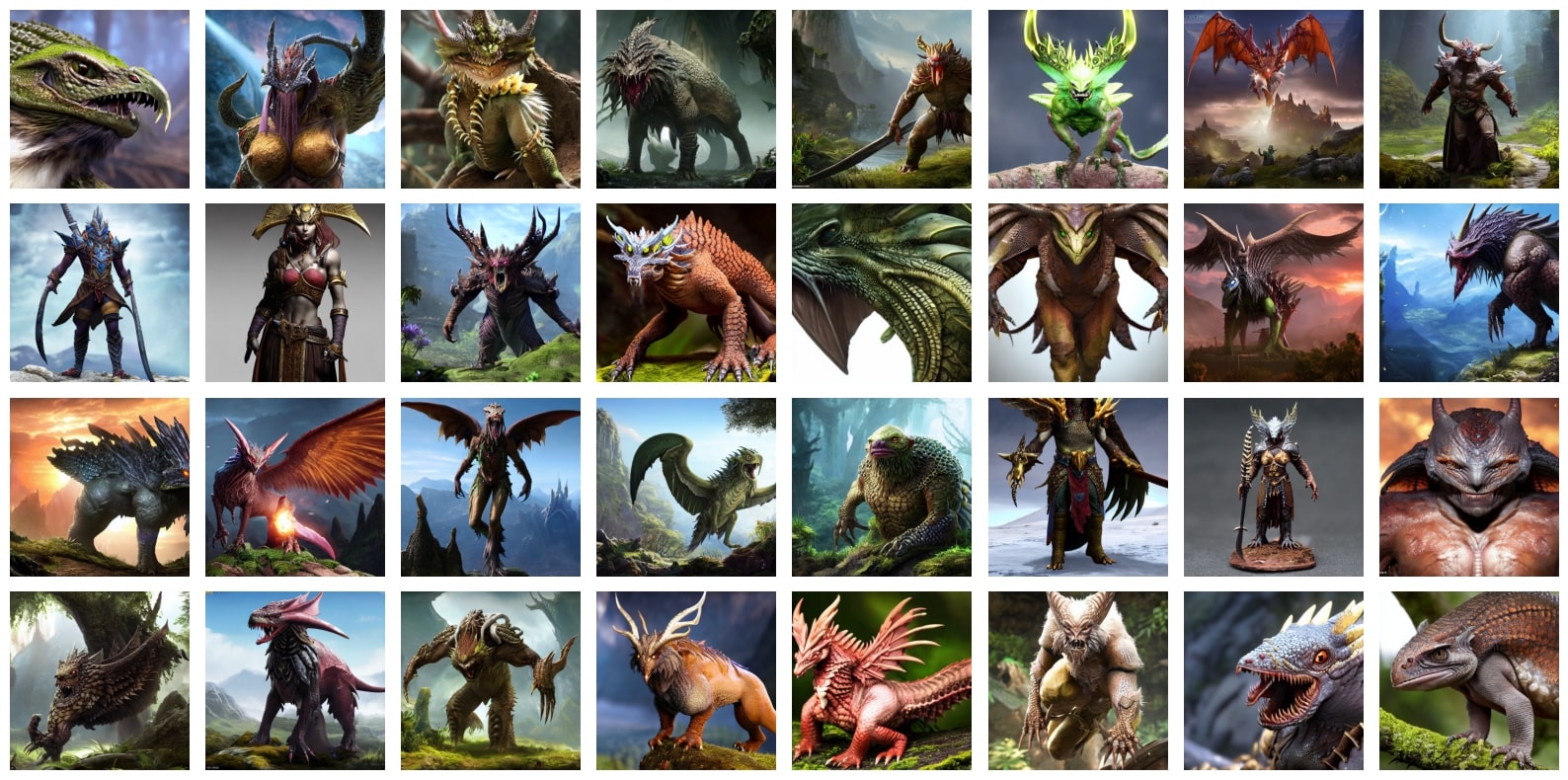}
    \par\vskip 0.5em  
    \includegraphics[width=0.8\linewidth]{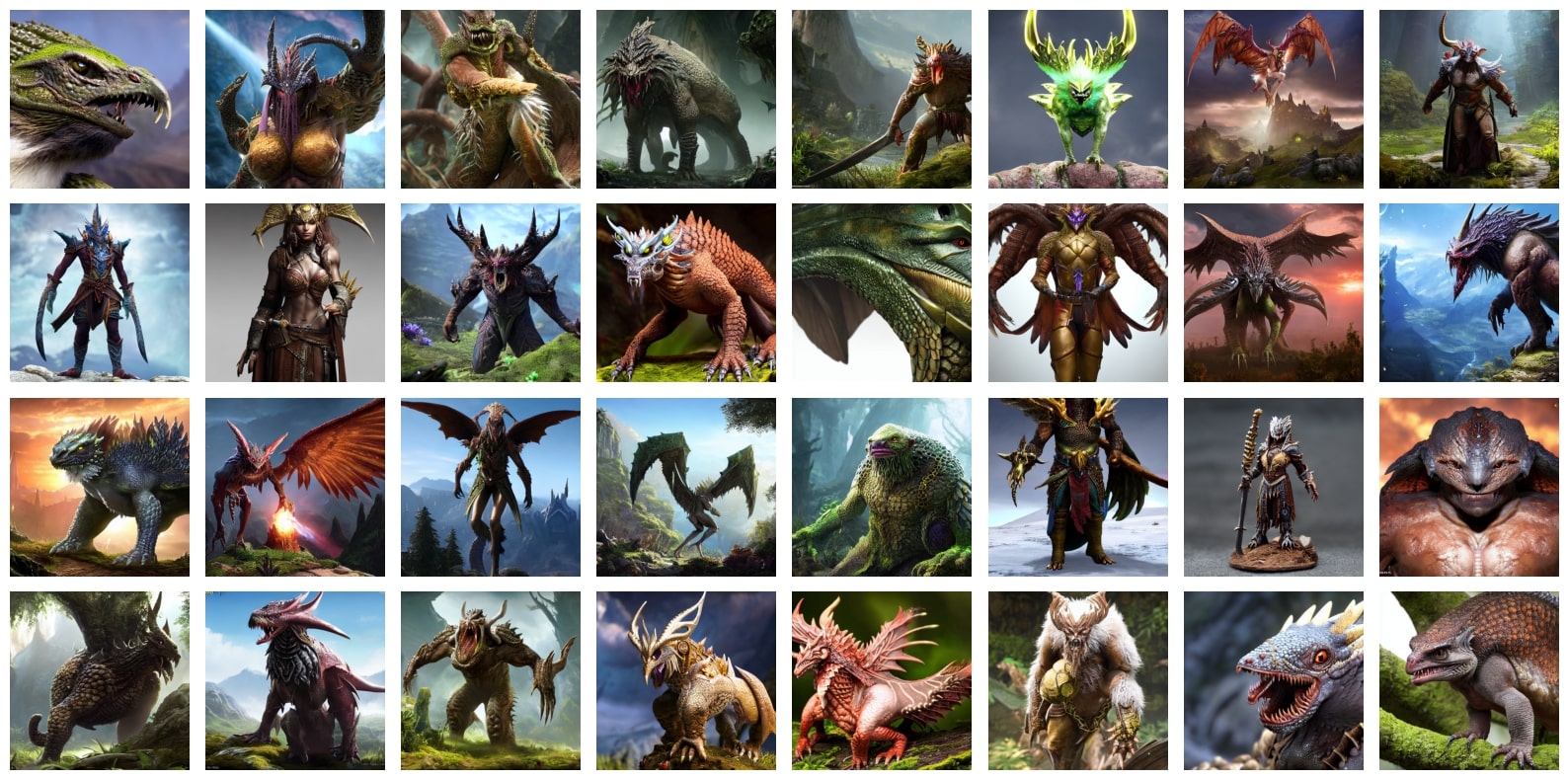}
    \caption{
    Top: Generated images using original model. Bottom: Generated images after applying SelfDebias. In the original set (top), only 30\% of the scenes depict humanoid-like creatures, whereas in the debiased set (bottom), this increases to 40\%, indicating a more balanced semantic distribution.} 

    \label{fig:fantasy_creature_grid}
\end{figure}

In the second example, we examine the model’s behavior when prompted with \textit{``a fantasy creature''}. Interestingly, the outputs tend to cluster into two broad semantic categories: one consisting of humanoid entities and the other dominated by beast-like forms resembling dinosaurs. We observe a bias toward the latter, with the model disproportionately producing dragon-like creatures while under-representing humanoid variants. SelfDebias successfully identifies and mitigates this imbalance. As illustrated in Fig.~\ref{fig:fantasy_creature_grid}, SelfDebias leads to a more semantically balanced distribution without requiring predefined concept axes.

\subsection{Generalization}

\textbf{Generalization to Imbalanced Targets}
Our method generalizes to arbitrary target distributions over discovered semantic clusters. This is especially useful for simulating specific environments, such as demographic skews in urban settings. Fig. 2 in supplementary material shows results. \\
\textbf{Generalization across image encoders: } To evaluate the independence of our debiasing method from the choice of image encoder, we additionally experiment with OpenCLIP's image encoder~\cite{cherti2023reproducible} for debiasing the gender attribute in the facial domain. As shown in Fig. 3 of supplementary material, our method successfully debiases the representations when using OpenCLIP. \\
\textbf{Generalization across model} To demonstrate the model-agnostic nature of SelfDebias, we evaluate it on an unconditional DDIM model trained on CelebA-HQ dataset~\cite{lee2020maskgan}. Despite the absence of text conditioning, our framework reduces demographic imbalance. We have shown qualitative results in Fig. 4 of supplementary. This confirms that our method generalizes effectively across architectures and training regimes. Unsupervised methods like UCE~\cite{gandikota2024unified} and TIME ~\cite{orgad2023editing}, cannot debias unconditional diffusion models because of their reliance on text tokens, whereas our method is generalizable to any diffusion model with an  UNet.

\section{Summary and Conclusion}
\label{sec:conclusion}

We propose SelfDebias, a fully unsupervised, framework for mitigating bias in diffusion models. SelfDebias guides generation by promoting uniformity over semantic clusters discovered in a latent space, without relying on attribute labels or external classifiers. 
Operating entirely in the  {\em h-space}, it generalizes across architectures that uses UNet as their backbone.
A limitation of our method is that discovered clusters may not always align with human-interpretable concepts. It can successfully debias only if those concepts are generated by the model. 
In conclusion, we hope that  SelfDebias contributes to the  pursuit of fair generation, especially in settings where supervision is limited but ethical generation is critical, and serves as a strong baseline for future unsupervised debiasing methods.
\section*{Acknowledgment} 
\label{sec:ack}
\textit{This work is partially funded by the Kotak IISc AI–ML Centre, IISc, whose support is gratefully acknowledged.}
{
    \small
    \bibliographystyle{ieeenat_fullname}
    \bibliography{main}
}

\end{document}


\begin{center}
    \Large\textbf{Supplementary Material} \\[1ex]
    \large \textbf{Unsupervised Self-debiasing of Text-to-Image Diffusion}\\
\end{center}

\vspace{1em}

\section{Training the Projector Network}
\label{sec:projector-training}

To map intermediate diffusion activations to semantic space, we use CLIP’s image encoder as a reference. For each timestep \( t \), the diffusion model yields an intermediate activation \( h_t \) at the U-Net bottleneck. We train a projector network \( g_\psi(h_t, t) \) to predict the CLIP embedding of the final generated image.

Each training sample consists of \( h_t \), the timestep \( t \), and three CLIP embeddings: one true and two perturbed positives, used to form contrastive training pairs. We optimize the projector using the NT-Xent loss and the Adam optimizer with a learning rate of \(10^{-4}\) over 30 epochs (batch size 256). Early stopping is based on validation cosine similarity.

\begin{figure}[H]
    \centering
    \includegraphics[width=0.5\linewidth]{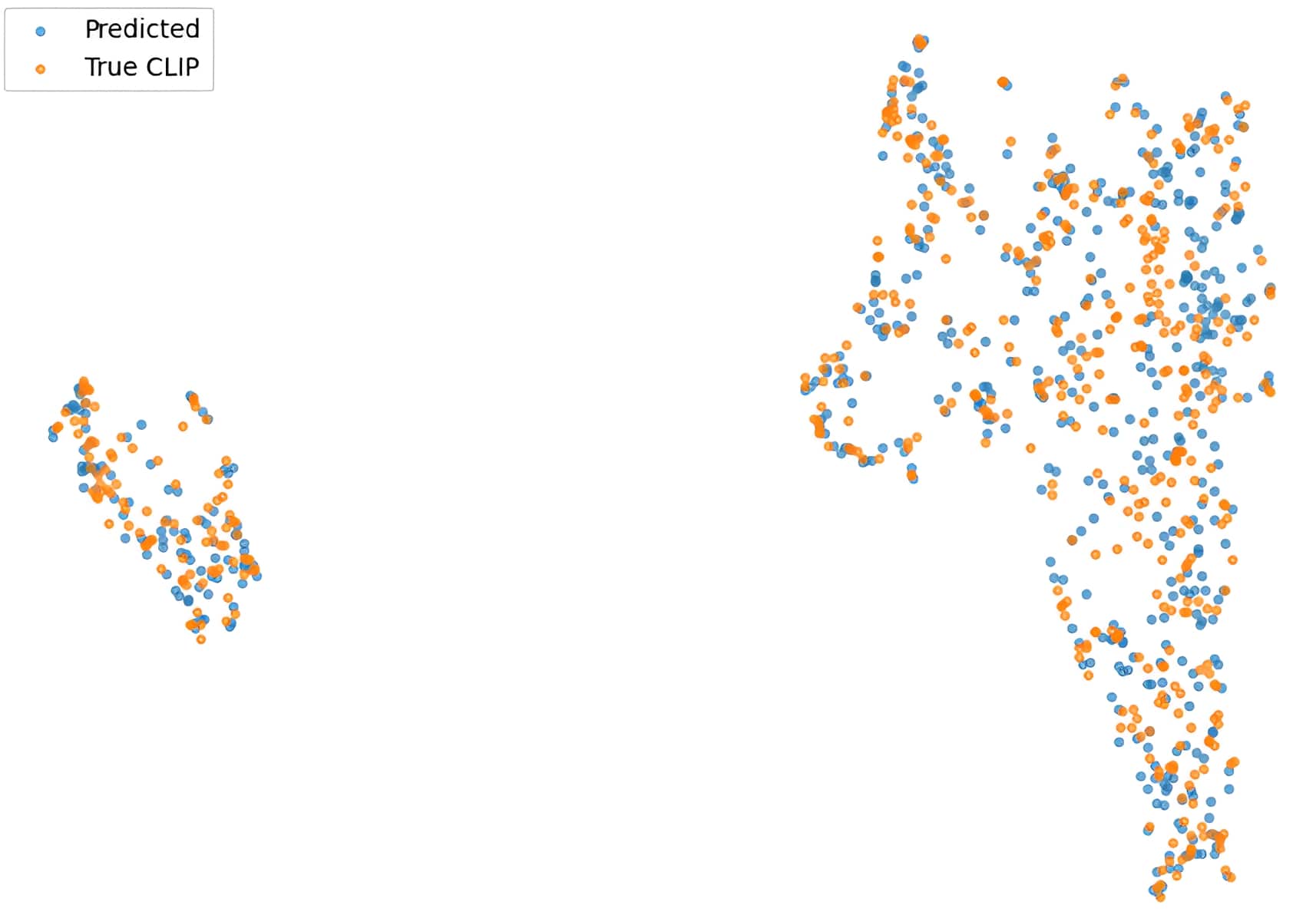}
    \caption{UMAP visualization showing alignment between original and predicted CLIP vectors for 'faces'. Cosine similarity = 0.9315.}
    \label{fig:proj_faces}
\end{figure}

Figure~\ref{fig:proj_faces} shows a UMAP projection, illustrating alignment between predicted and ground-truth CLIP embeddings.

\section{Imbalanced Target Distributions}

\begin{figure}[H]
    \centering
    \includegraphics[width=0.5\linewidth]{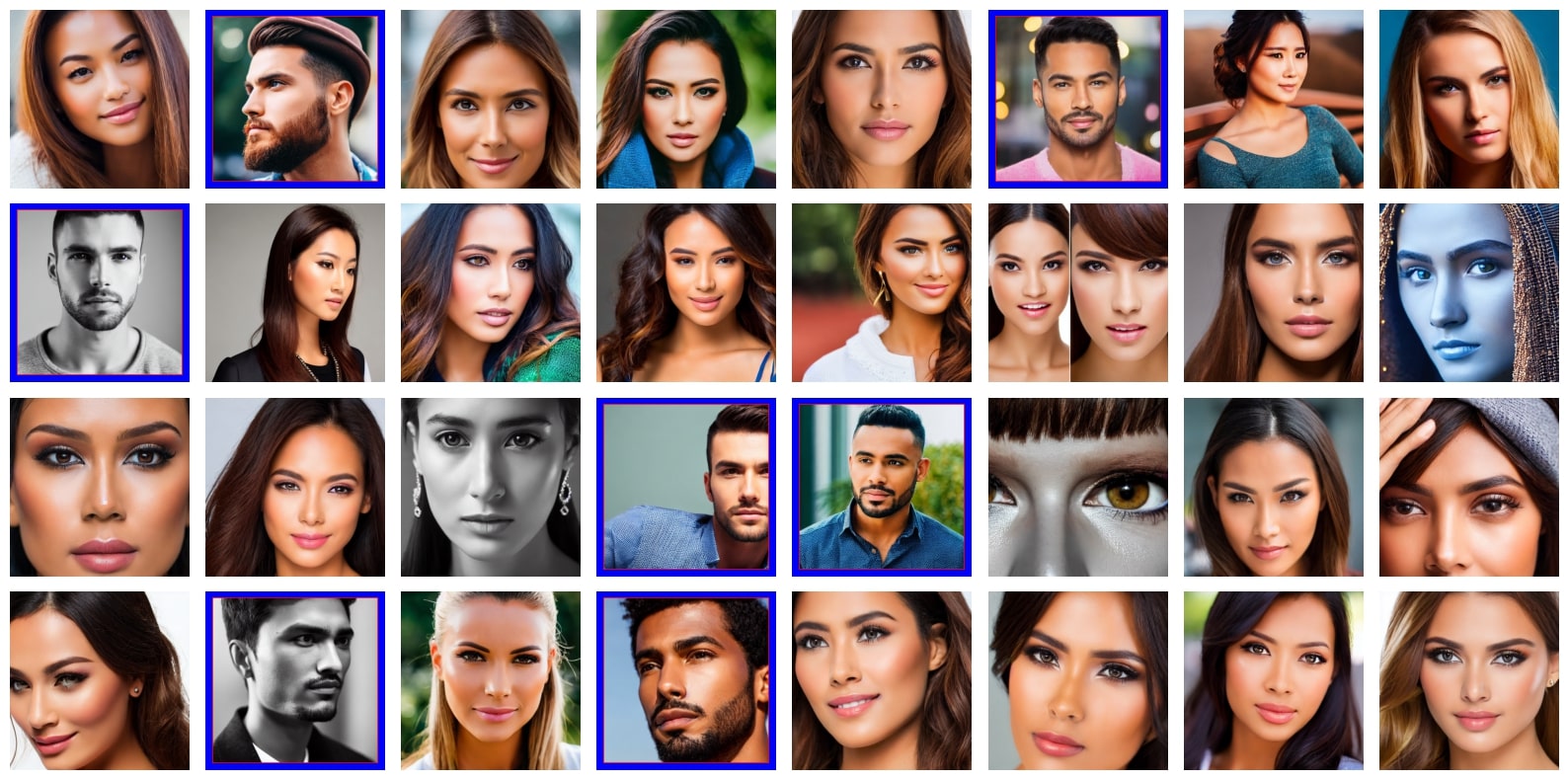}
    \par\vskip 1em
    \includegraphics[width=0.5\linewidth]{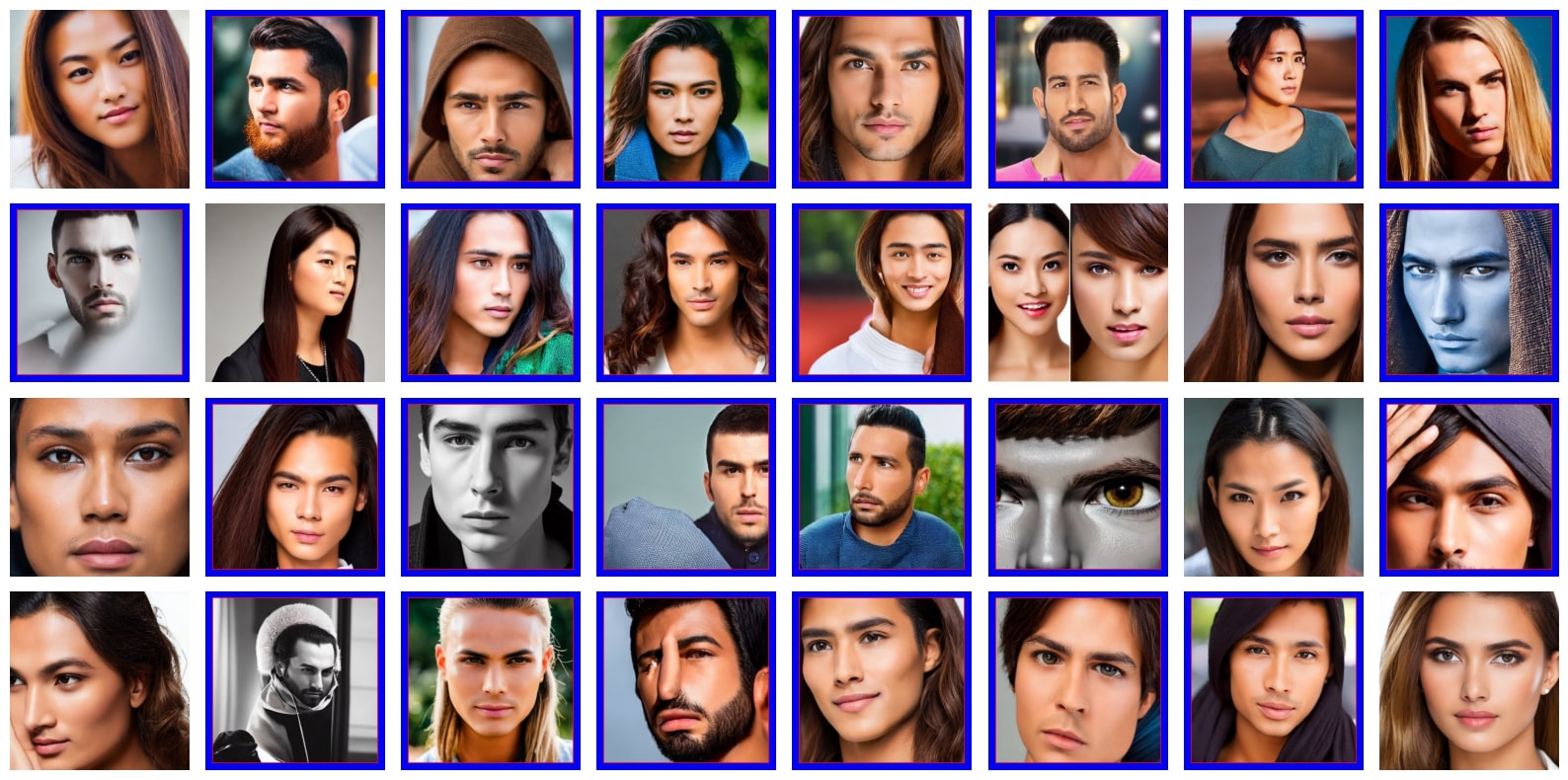}
    \caption{Imbalanced Target Debiasing. Top: Original (unbiased) generation. Bottom: Debiased output using Self-Debias steered toward 70\% male subjects.}
    \label{fig:imbalanced_grid}
\end{figure}

Our method generalizes to arbitrary target distributions over discovered semantic clusters. This is especially useful for simulating specific environments, such as demographic skews in urban settings. Figure~\ref{fig:imbalanced_grid} shows results for a 70:30 male:female target.

\section{Using OpenCLIP Image Encoder}

\begin{figure}[H]
    \centering
    \includegraphics[width=0.5\linewidth]{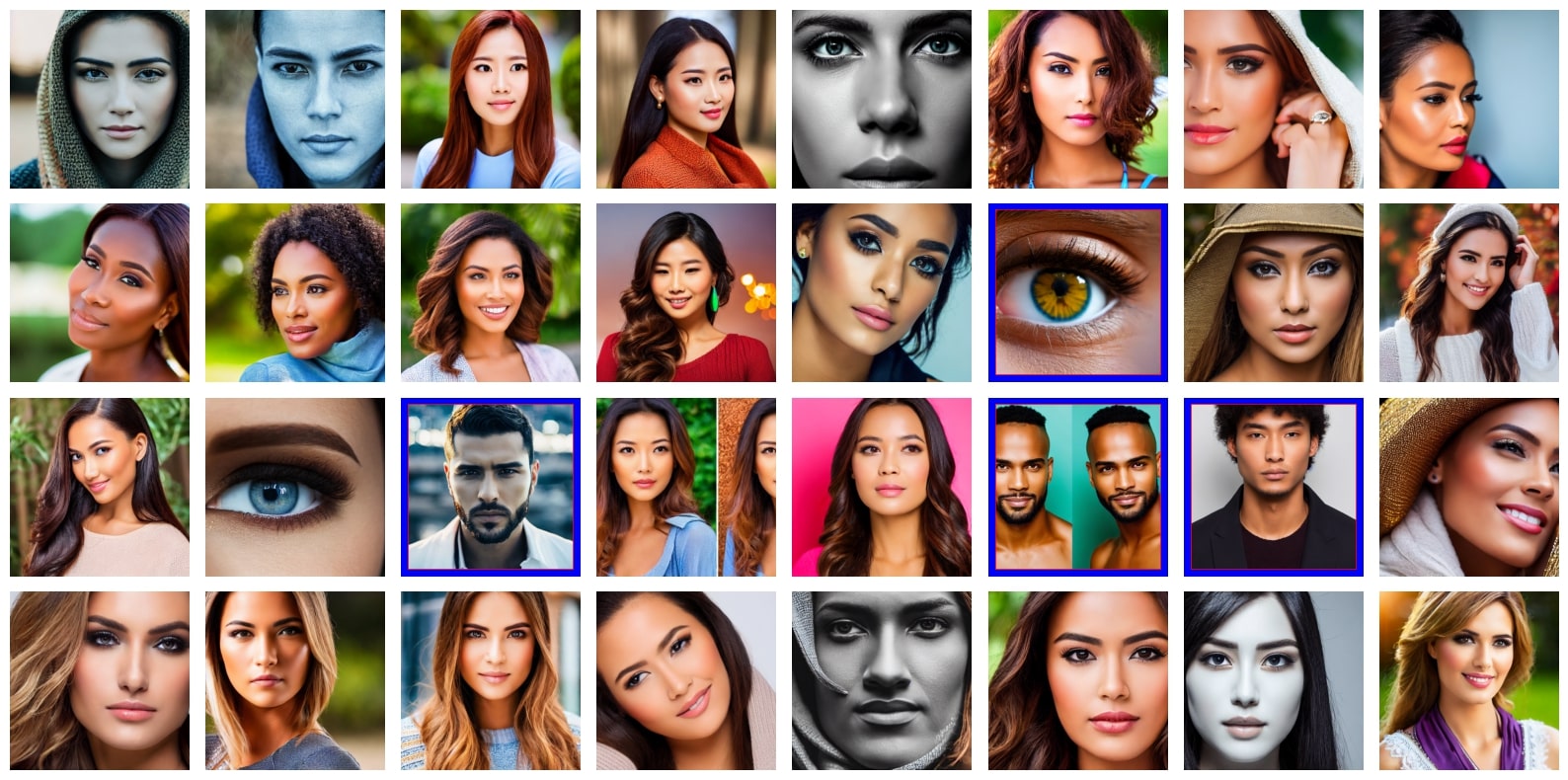}
    \par\vskip 1em
    \includegraphics[width=0.5\linewidth]{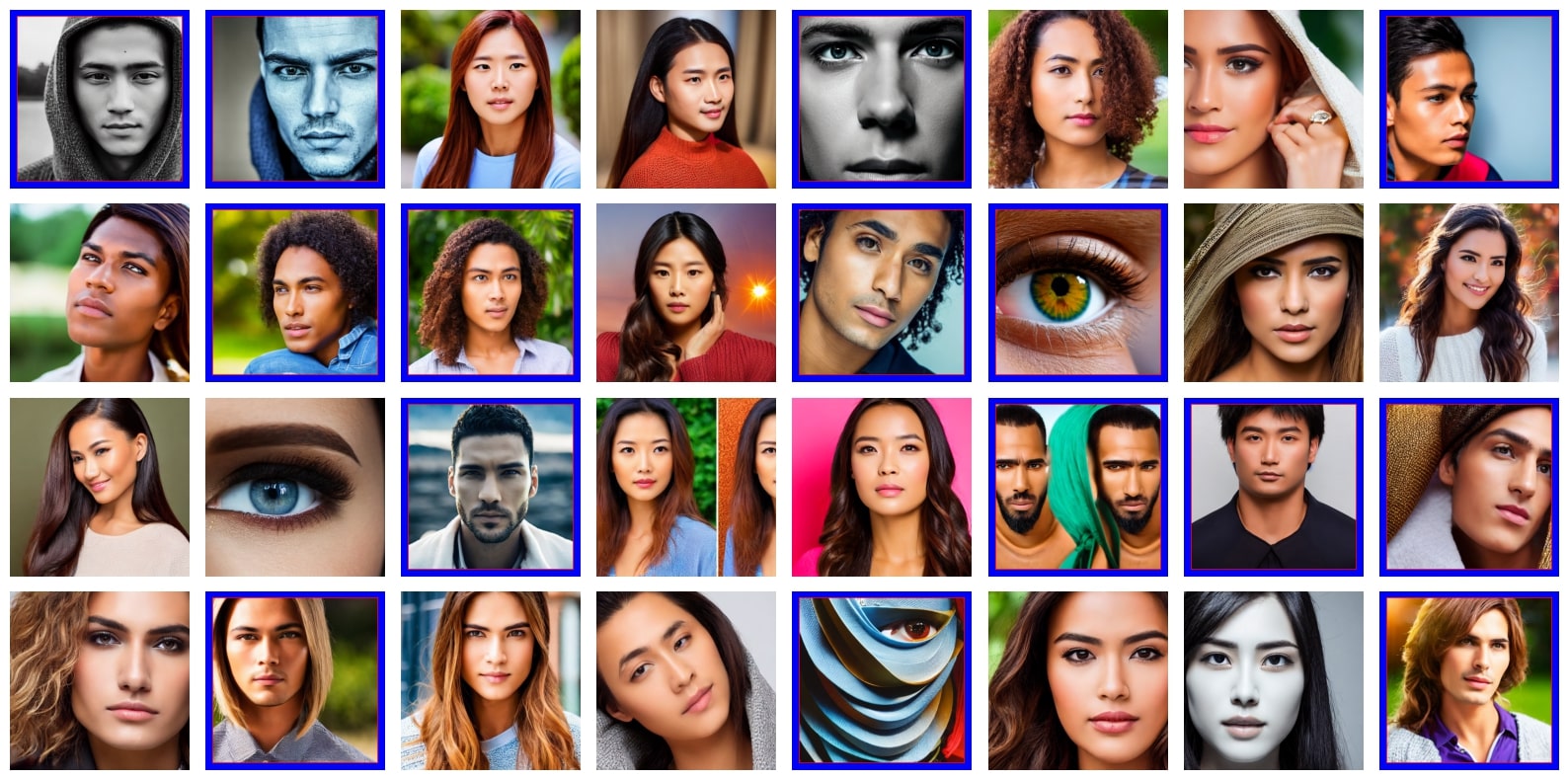}
    \caption{OpenCLIP Debiasing. Top: Without debiasing. Bottom: Debiased output shows improved male representation (14 male-presenting subjects vs 3 originally).}
    \label{fig:openclip_results}
\end{figure}

We also evaluate performance when using OpenCLIP instead of CLIP. As shown in Figure~\ref{fig:openclip_results}, our method remains effective at debiasing, yielding better gender balance even with a different encoder.

\section{Unconditional Model Evaluation}

\begin{figure}[H]
    \centering
    \includegraphics[width=1\linewidth]{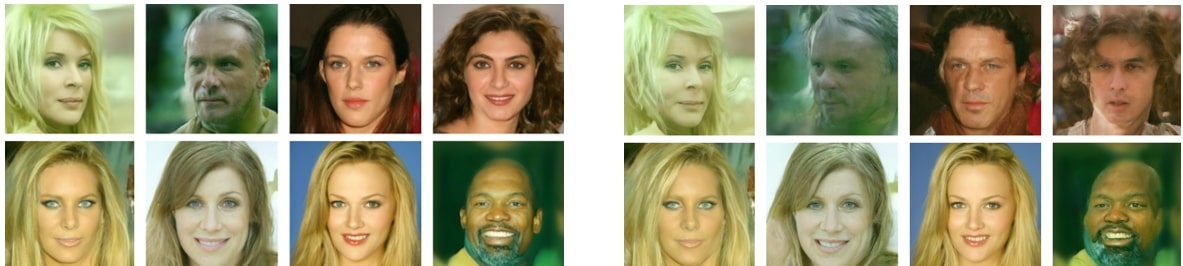}
    \caption{DDIM Debiasing — Gender: Left = baseline, Right = debiased. Debiased output has equal male/female ratio (4 each) versus only 2 males in original. Nearest unsupervised baseline method, \textbf{UCE} couldn't do unconditional debiasing because of its reliance on text embeddings.}
    \label{fig:ddim_faces}
\end{figure}

Figure~\ref{fig:ddim_faces} shows qualitative diversity confirming the debiasing of unconditional diffusion.

\section{Applicability to Transformer-based Backbones}
\label{sec:transformer_backbones}

While our main experiments focus on diffusion models with U-Net backbones, our framework is not restricted to this architecture. The only requirement is access to intermediate semantic representations during the sampling process. For transformer-based diffusion and rectified-flow models such as FLUX and HiDream, the natural analogue of our \textit{h}-space is the sequence of hidden states corresponding to image tokens at denoising step \(t\) and transformer layer \(l\).

These token-level hidden states contain rich semantic information, similar in spirit to the bottleneck activations in U-Net models. In principle, our Semantic Projection Module (Sec. 3.1 in the main paper) can be trained to align these hidden states with a semantic embedding space (e.g., CLIP or OpenCLIP). Subsequently, the Semantic Mode Discovery and Self-debiasing Modules (Secs. 3.2 and 3.3 in the main paper) can be applied in exactly the same manner, using the projected transformer features instead of projected U-Net features.

A few architecture-specific considerations arise:
\begin{itemize}
    \item \textbf{Hook points:} In transformers, intermediate hidden states are available at every layer. Selecting which layer(s) to hook is an open design choice. Earlier layers may encode low-level structural information, while deeper layers capture high-level semantics.
    \item \textbf{Sequence length:} Transformer backbones operate over long token sequences, leading to higher memory and compute costs. Pooling strategies (e.g., mean pooling, attention pooling, or downsampling subsets of tokens) can reduce dimensionality while retaining semantic fidelity.
    \item \textbf{Token granularity:} Unlike U-Net activations, token embeddings correspond to discrete spatial patches or latent codes. The choice of token granularity (e.g., patch size or latent resolution) will impact both semantic separability and computational efficiency.
\end{itemize}

While these considerations require empirical exploration, they are orthogonal to our main contribution. Our framework is fully compatible with transformer-based architectures, and extending SelfDebias to such backbones is a promising direction for future work.

\section{Ablation on Number of Gradient Update Steps}
\label{sec:grad_steps}

In our main framework, we apply a single gradient update to the \emph{h}-space per denoising step (Eq.~(8)) for efficiency. To study the effect of taking multiple gradient updates, we perform an ablation with 1, 2, 3, and 5 inner updates per denoising step. 

\begin{table}[h]
\centering
\small
\begin{tabular}{lcccc}
\toprule
\# Steps & 1 & 2 & 3 & 5 \\
\midrule
FD $\downarrow$  & 0.009  & 0.0088 & 0.0087 & 0.0087 \\
FID $\downarrow$ & 70.52  & 70.85  & 77.53  & 88.32 \\
\bottomrule
\end{tabular}
\caption{Ablation on the number of gradient updates per denoising step. FD improves marginally with more steps, while FID degrades as the number of updates increases. This is for gender attribute for faces data.}
\end{table}

We observe that Fairness Discrepancy (FD) improves slightly as the number of updates increases, indicating marginal gains in bias mitigation. However, FID worsens substantially beyond two steps. This behavior can be explained as follows: each additional update step applies stronger corrective pressure toward the uniform cluster distribution, which reduces demographic skew but simultaneously pushes the denoising trajectory farther away from the model’s natural manifold. As a result, semantic alignment improves (lower FD) but visual fidelity degrades (higher FID). Beyond two steps, this over-correction leads to overshooting and accumulation of artifacts. Therefore, our default choice of one gradient update per denoising step achieves the best trade-off between fairness and image quality, while also remaining computationally efficient.

\section{Runtime Analysis}
\label{sec:runtime}

We report a runtime comparison between our framework and the vanilla diffusion model in Table~\ref{tab:runtime_a5000}. Our approach introduces two one-time offline stages: (i) training the projector network and (ii) identifying semantic modes via clustering. These costs are incurred once per prompt family, after which the centroids can be reused across related prompts (as demonstrated in Sec.~4.3 where centroids derived from faces are reused for occupation-based prompts). The per-sample overhead during inference is modest: we apply lightweight gradient updates in \emph{h}-space without backpropagating through the full U-Net.

\begin{table}[t]
\centering
\caption{Runtime comparison on an RTX A5000 GPU. Preprocessing is a one-time offline cost; scope indicates whether it is per attribute or per prompt family.}
\label{tab:runtime_a5000}
\small
\setlength{\tabcolsep}{6pt}
\renewcommand{\arraystretch}{1.0}
\begin{tabular}{lcc}
\toprule
\textbf{Method} & \textbf{One-time preprocessing (scope)} & \textbf{Sampling (per image)} \\
\midrule
Vanilla Stable Diffusion & None & 1.78 s \\
H-Guidance & 2.93 h \,/ attribute & 2.85 s \\
Self-Discovery & 4.25 h \,/ attribute & 3.02 s \\
\textbf{Ours} & 3.63 h \,/ prompt family & 3.34 s \\
\bottomrule
\end{tabular}
\end{table}

Overall, the results show that while our method introduces a one-time offline cost, the inference-time overhead remains practical, roughly 1.87X sampling time relative to vanilla diffusion. Crucially, because centroids can be reused across prompt families, our framework enables real-time debiasing applications once this offline stage has been performed.

\section{Hyper-parameter Values and Ablation}
\label{sec:hyperparams}

We use a small, fixed set of hyper-parameters across all experiments, without tuning them within a prompt family.
\begin{itemize}
    \item \textbf{Stage-1 cluster count $k$:} chosen via a silhouette-score sweep on projected features (typically 2--5 across prompt families).
    \item \textbf{Stage-2 refinement:} recursive spectral splits with a default $d_{\max}=3$. We set $s_{\min}=0.05N$ (i.e., a minimum leaf size of 5\% of the dataset) to prevent very small, impure leaves whose centroids would be unrepresentative. Because $d_{\max}$ controls \emph{how deep} we recurse, it is more sensitive at shallow depths (coarse vs.\ mid-level partitions). Beyond a certain depth, however, $s_{\min}$ blocks further splits, so increasing $d_{\max}$ has diminishing effect.
    \item \textbf{Temperature $\alpha$:} fixed globally at $\alpha=8$, controlling the sharpness of the soft assignment in Eq.~(3).
\end{itemize}

\paragraph{Sensitivity of $\alpha$ (gender).} Table~\ref{tab:ablation_alpha_gender} shows a mild trend: $\alpha{=}6$ slightly lowers FD, but at a small cost to FID; $\alpha{=}8$ attains the best overall balance (lower FID and stable FD), and we use it as default.

\begin{table}[h]
\centering
\small
\setlength{\tabcolsep}{8pt}
\begin{tabular}{lcccccc}
\toprule
$\alpha$ & 4 & 6 & 8 & 10 & 12 & 16 \\
\midrule
FD $\downarrow$  & 0.017 & \textbf{0.014} & 0.015 & 0.015 & 0.016 & 0.017 \\
FID $\downarrow$ & 87.53 & 87.28 & \textbf{87.08} & 87.33 & 87.72 & 88.19 \\
\bottomrule
\end{tabular}
\caption{Ablation on $\alpha$ for \textbf{gender}. While $\alpha{=}6$ slightly lowers FD, $\alpha{=}8$ offers the best \emph{balance} (lower FID and stability), so we adopt $\alpha{=}8$ as default.}
\label{tab:ablation_alpha_gender}
\end{table}

\paragraph{Sensitivity of $d_{\max}$ (race).} Table~\ref{tab:ablation_dmax_race} highlights that deeper recursion up to $d_{\max}{=}3$ is beneficial (coarse $\rightarrow$ mid-level structure). Pushing deeper to $d_{\max}{=}4$ starts to over-fragment, slightly hurting FD/FID. From $d_{\max}{=}4$ to $5$, changes are small; for $d_{\max}>5$, we observe \emph{no} changes because $s_{\min}$ prevents further splits, so additional depth is effectively inert.

\begin{table}[h]
\centering
\small
\setlength{\tabcolsep}{8pt}
\begin{tabular}{lcccc}
\toprule
$d_{\max}$ & 2 & 3 & 4 & 5 \\
\midrule
FD $\downarrow$  & 0.239 & \textbf{0.237} & 0.258 & 0.253 \\
FID $\downarrow$ & 87.82 & \textbf{87.08} & 87.63 & 87.85 \\
\bottomrule
\end{tabular}
\caption{Ablation on $d_{\max}$ for \textbf{race}. Early increases in depth are meaningful; beyond $d_{\max}{=}4$, $s_{\min}$ blocks most additional partitions, so $d_{\max}{=}5$ yields only minor changes, and $d_{\max}{>}5$ has no effect. Default $d_{\max}{=}3$ achieves the best trade-off.}
\label{tab:ablation_dmax_race}
\end{table}

Overall, FD/FID are robust around the defaults, with $d_{\max}$ mattering primarily at shallow depths and quickly saturating once $s_{\min}$ halts further splits. This supports using a single configuration ($\alpha{=}8$, $d_{\max}{=}3$) across a prompt family without per-attribute tuning.

\section{Two-Stage Clustering}
\label{sec:two_stage_rationale}

CLIP space exhibits a few \emph{dominant} axes with finer variation \emph{within} those groups, so we use a two-stage procedure that matches this structure.

\textbf{Stage-1 (coarse split).} Choose $k$ via the silhouette criterion to capture the dominant partition.

\textbf{Stage-2 (local refinement).} Recursively refine \emph{within} each Stage-1 cluster by spectral clustering, gated by $(s_{\min}, d_{\max})$, to reveal fine-grained variants while preserving the coarse split. Any Stage-1 cluster with $|c|\!\ge\! s_{\min}$ is eligible for refinement.

\begin{figure}[t]
\centering

\begin{subfigure}[t]{0.485\linewidth}
    \includegraphics[width=\linewidth]{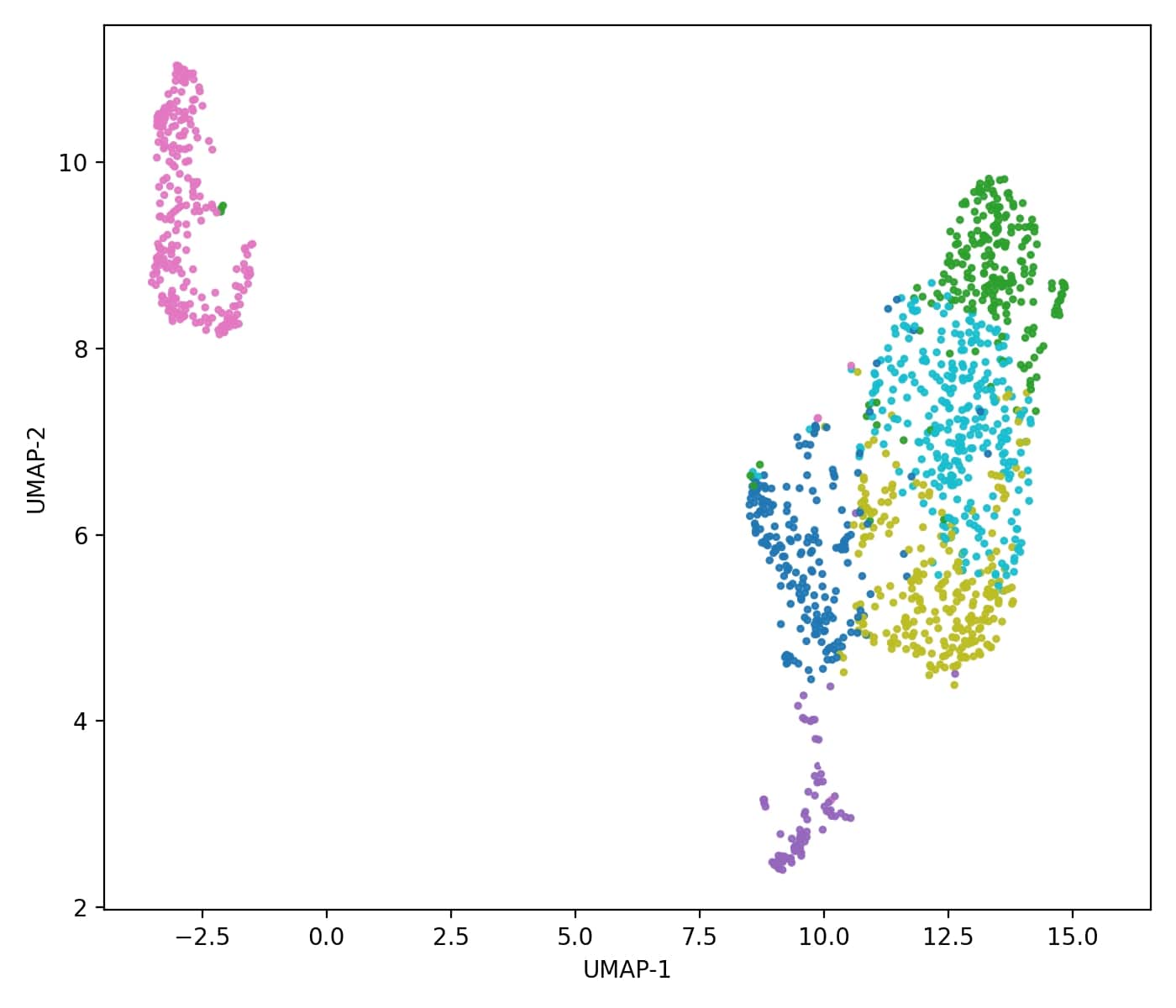}
    \caption{Flat clustering ($k{=}6$). All splits land in the larger coarse group; the smaller group remains unsplit.}
\end{subfigure}\hfill
\begin{subfigure}[t]{0.485\linewidth}
    \includegraphics[width=\linewidth]{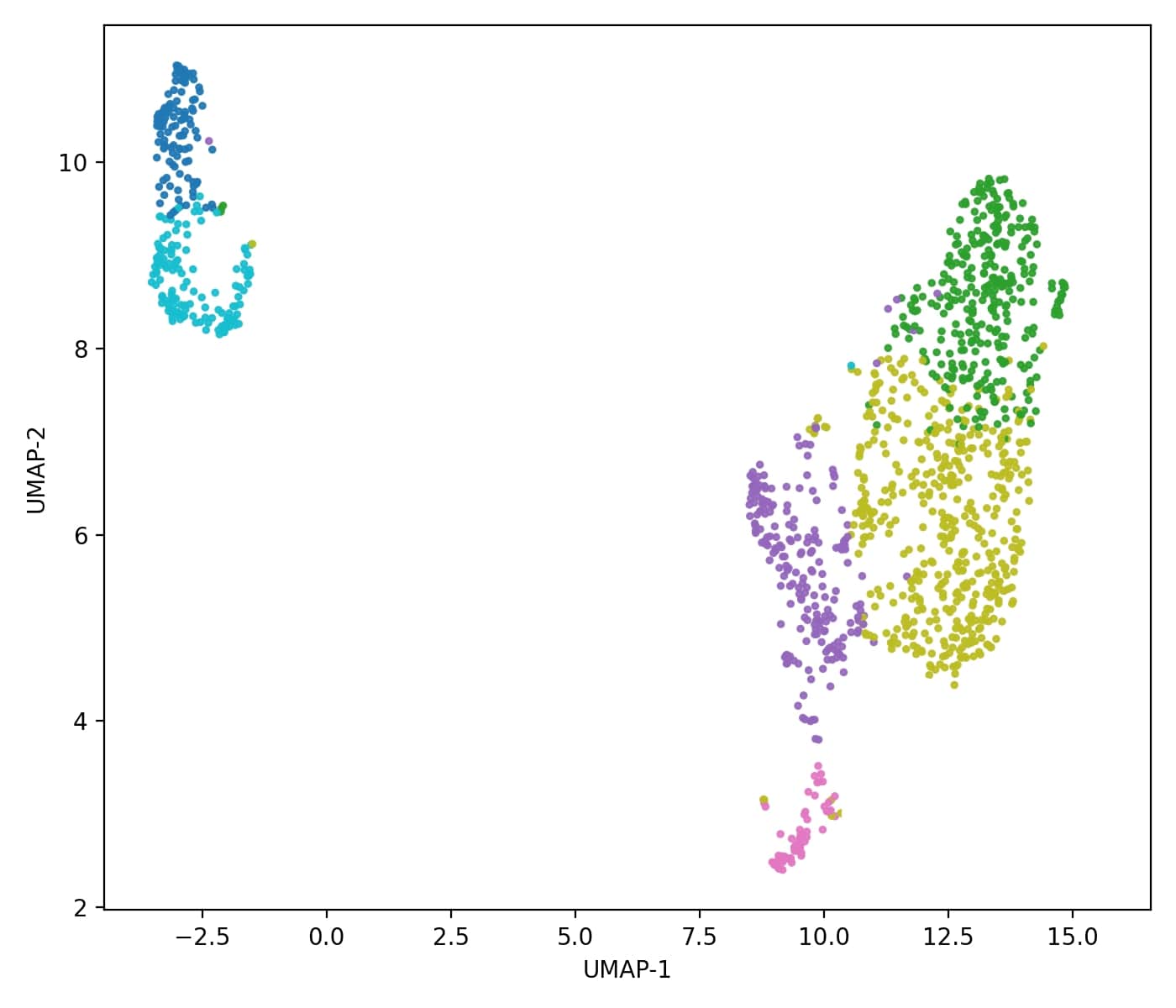}
    \caption{Two-stage (ours, 6 leaves). Stage-1 isolates coarse groups; Stage-2 refines within groups ($|c|\!\ge\! s_{\min}$).}
\end{subfigure}

\vspace{0.35em}

\begin{subfigure}[t]{0.485\linewidth}
    \includegraphics[width=\linewidth]{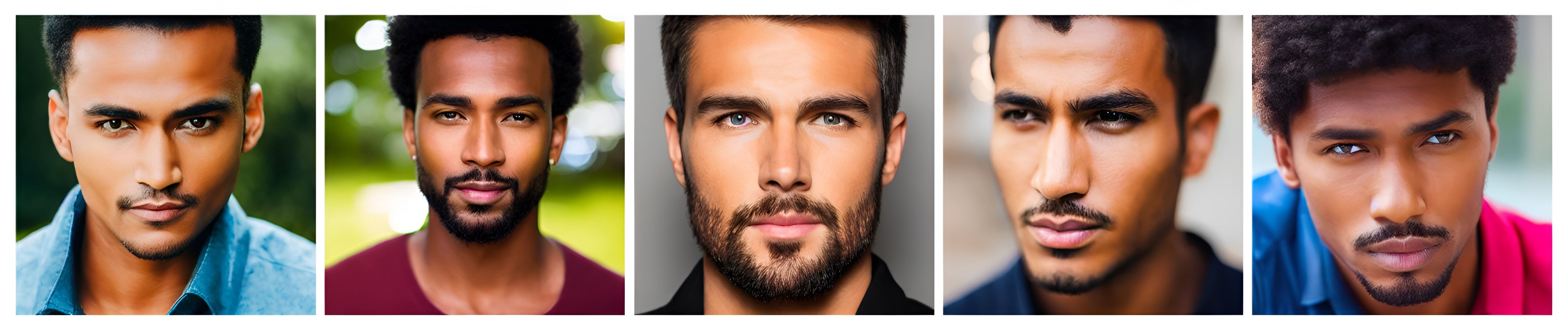}
    \caption{Sample images closest to the centroid of the pink cluster in (a); clearly show a mixture of races.}
\end{subfigure}\hfill
\begin{subfigure}[t]{0.485\linewidth}
    \includegraphics[width=\linewidth]{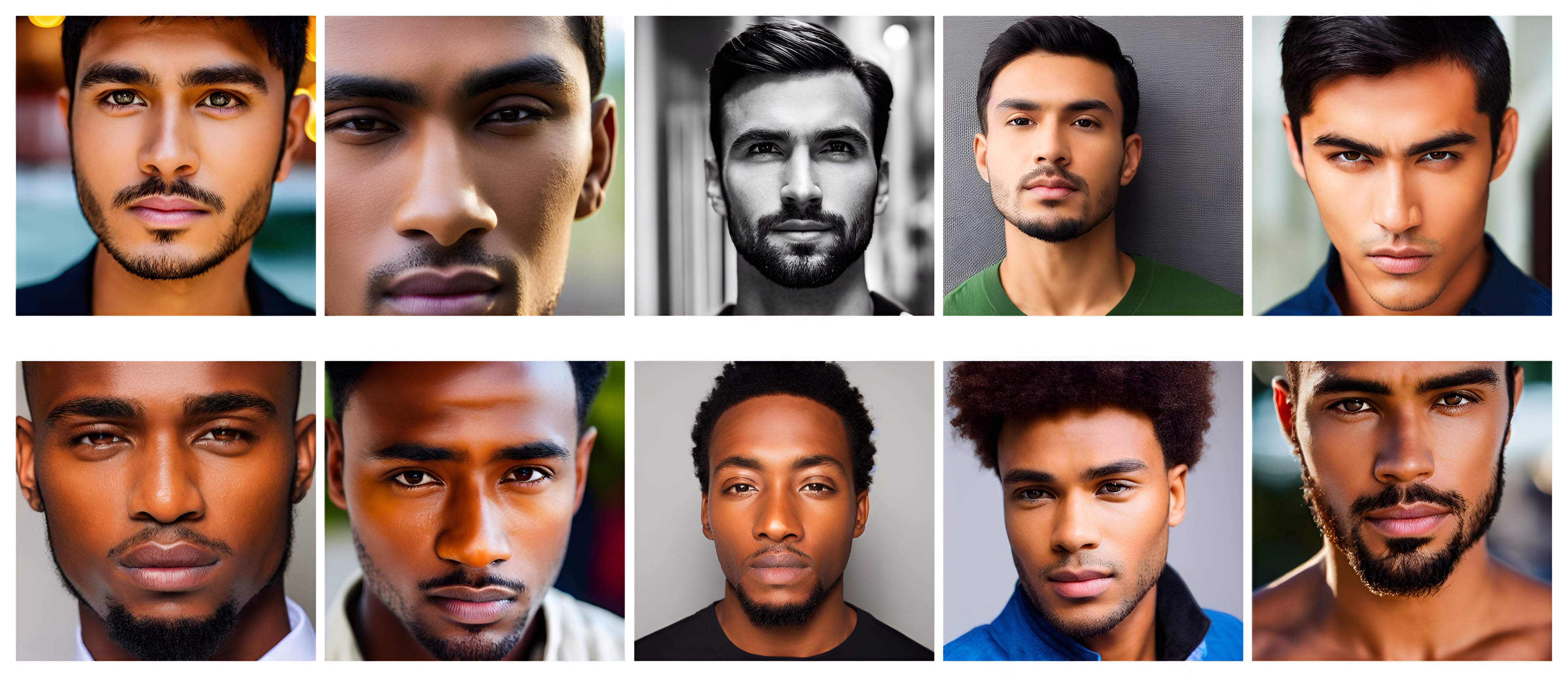}
    \caption{Sample images closest to the centroids of the blue and sky-blue clusters in (b) (top: blue, bottom: sky-blue); clearly show separation by race.}
\end{subfigure}

\caption{Human face clustering and race-based selections in projected CLIP space (UMAP) and zero-shot filtering. (a,b) compare flat vs. hierarchical clustering with the same total leaves ($k{=}6$), showing control over \emph{where} splits occur. (c) shows a single mixed row across races; (d) shows two separated rows by race, with non-overlapping images.}
\label{fig:faces_flat_vs_hier_k6}
\end{figure}

\textbf{Why not a flat global $k$?} With a single global $k$, one cannot control \emph{where} splits land: even at the same leaf count (e.g., $k\!=\!6$), the optimizer may allocate most splits to the larger coarse group and leave the smaller group unsplit. Our locally gated refinement ensures both large and smaller coarse groups are split when warranted, without retuning $k$. Figure \ref{fig:faces_flat_vs_hier_k6} shows such a scenario.

\section{Debiasing Non Societal Biases}
\label{sec:debiasing_vague}
\begin{figure}[H]
    \centering
    \includegraphics[width=0.5\linewidth]{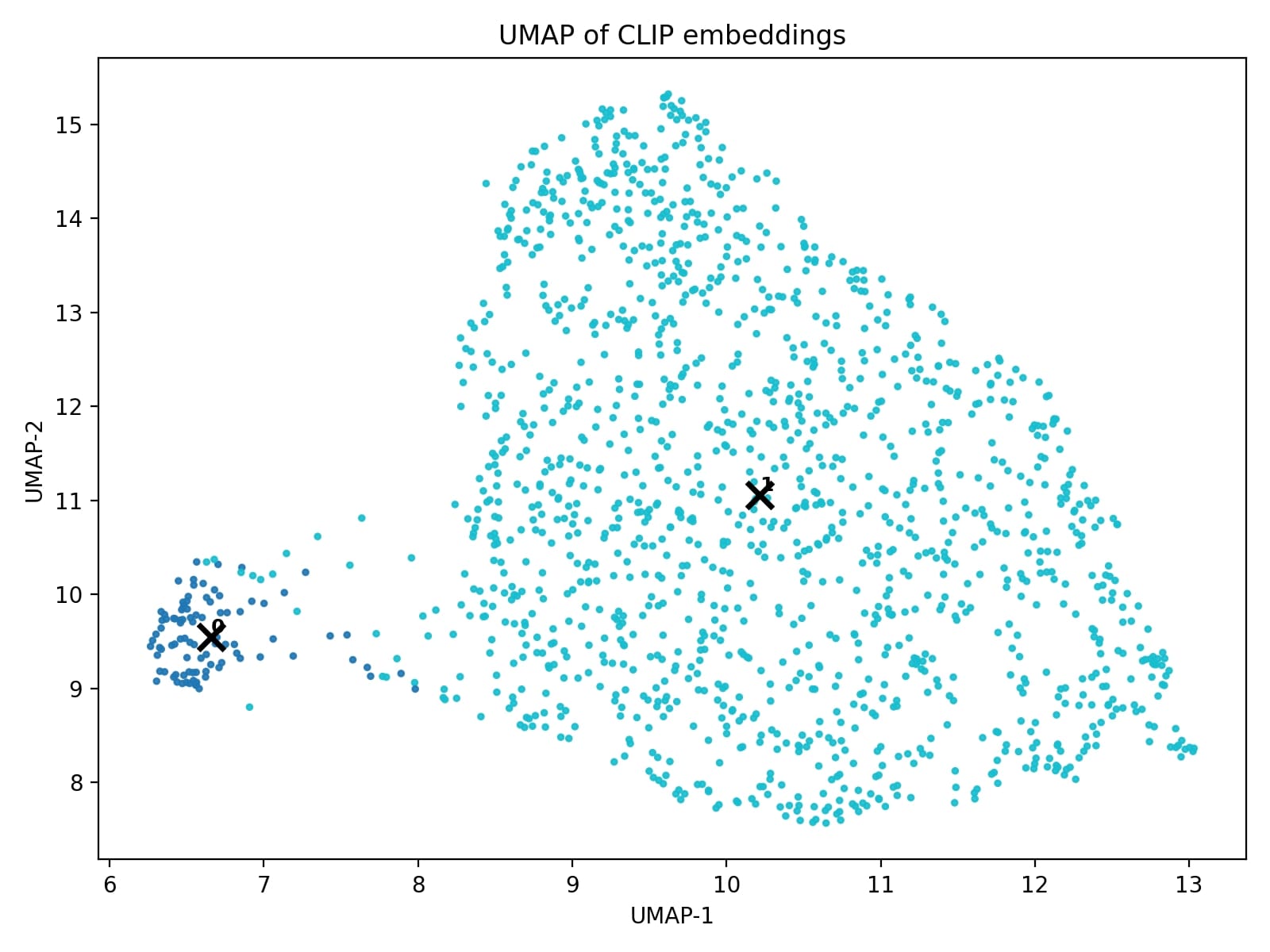}
    \caption{CLIP clusters formed when prompted with \textit{'a photo of food on a table'}.}
    \label{fig:clip_food_clusters}
\end{figure}

\begin{figure}[H]
    \centering
    \includegraphics[width=0.5\linewidth]{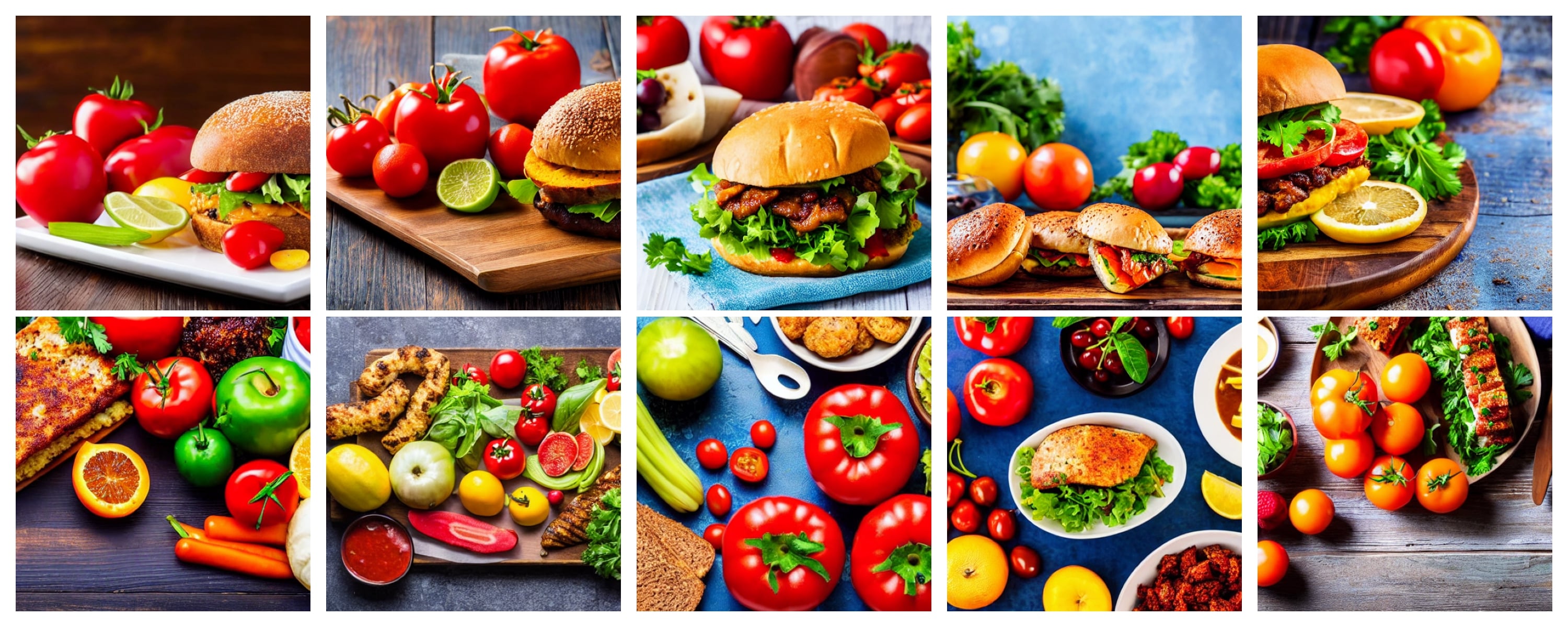}
    \caption{Representative images from each of the CLIP clusters, each row corresponds to a cluster.}
    \label{fig:rep_food_images}
\end{figure}

\begin{figure}[H]
    \centering
    \includegraphics[width=0.5\linewidth]{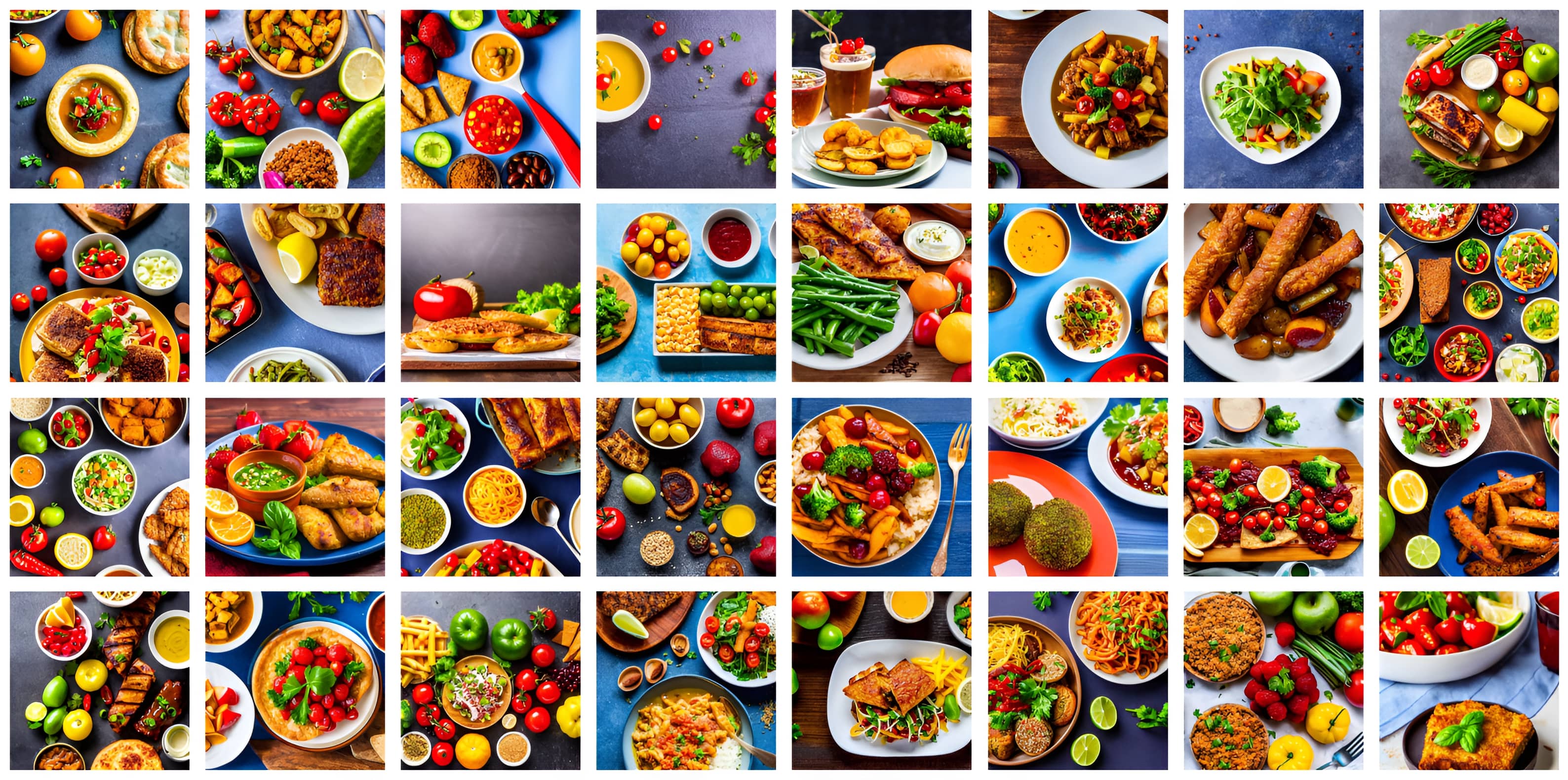}
    \par\vskip 0.5em
    \includegraphics[width=0.5\linewidth]{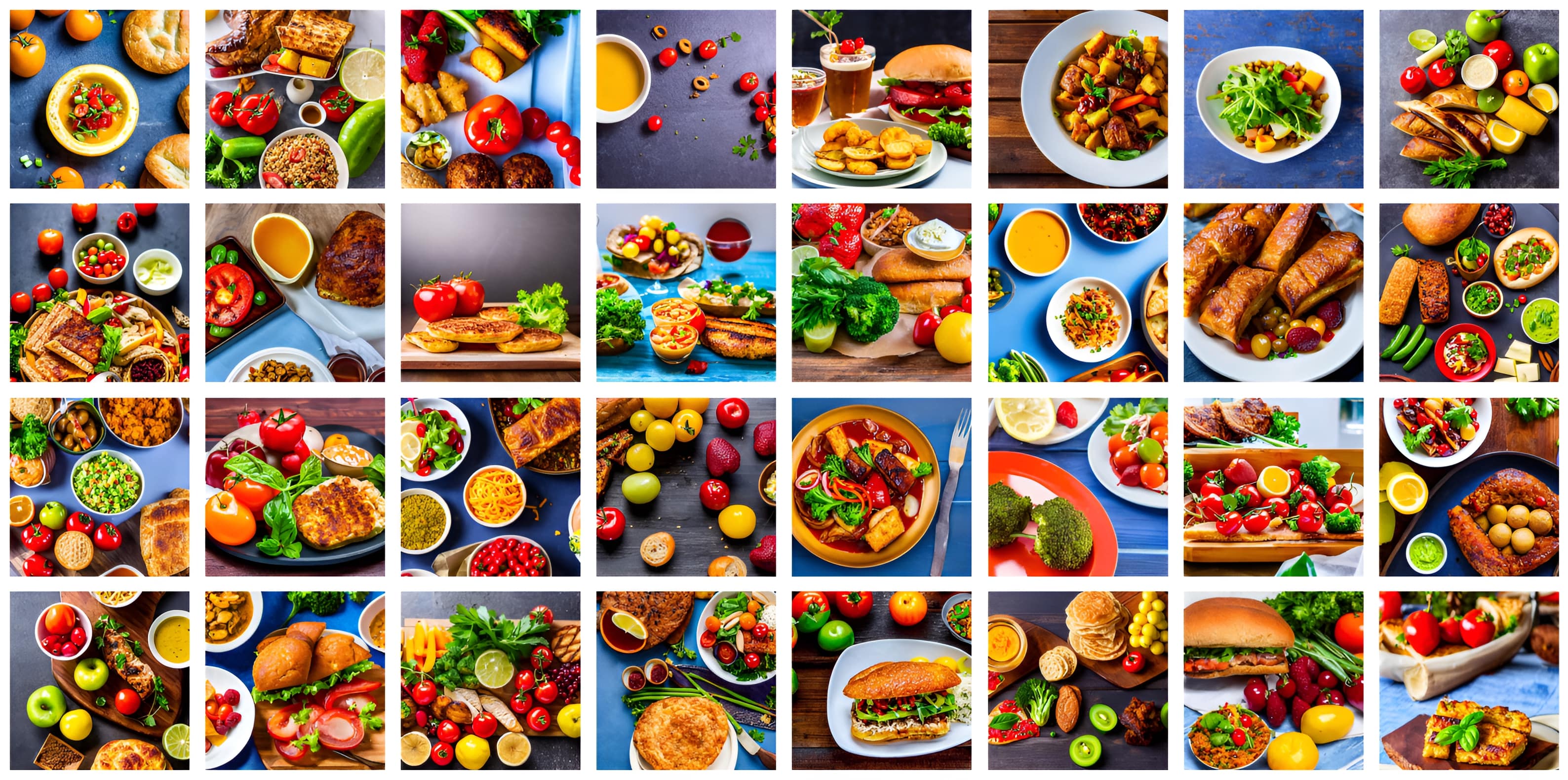}
    \caption{
        Top: Generated images using vanilla Stable Diffusion 1.5. \color{black}
        Bottom: Generated images using SelfDebias. The debiased set contains 5 burgers, as opposed to only 1 using the original model.
    }
    \label{fig:stacked_foods}
\end{figure}

Our method works not only for societal biases but also for biases while generating niche or abstract concepts as shown in Sec. 4.4 of the main paper. We here provide one more example of a non-societal bias where we debias images generated when the SD model is prompted with the prompt \textit{'a photo of food on a table'}. On seeing the clusters formed, it is evident that larger cluster belongs to images containing only veggies and smaller cluster corresponds to images with burgers. We have shown the clusters in Fig. ~\ref{fig:clip_food_clusters}. We have debiased using the centroids found using the SelfDebias method and the result is shown in Fig. \ref{fig:stacked_foods}. We have also shown in Fig. \ref{fig:rep_food_images}, representative images from each cluster, randomly sampled. Since the number of available images in the smaller cluster is limited, and SelfDebias primarily shifts images near the decision boundary to the opposite cluster, we were unable to achieve a truly uniform distribution without compromising image quality. But we are able to decrease the bias as shown in the Fig. \ref{fig:stacked_foods}.

